\documentclass{article}
\usepackage{spconf,amsmath,graphicx,algorithm,algpseudocode,amsmath,amsfonts,amsthm,amssymb,amsopn,booktabs,rotating,multirow,hyperref}
\usepackage[skip=0pt]{caption}
\newcommand{\mat}[1]{\boldsymbol{#1}} 
\newcommand{\norm}[1]{\left\|#1\right\|} 
\newcommand{\inner}[2]{#1\cdot #2} 

\title{A Fully Convolutional Tri-branch Network (FCTN) for Domain Adaptation}
\name{Junting Zhang, Chen Liang and C.-C. Jay Kuo}
\address{University of Southern California, Los Angeles, CA, USA\\
\small{\texttt{\{\href{mailto:juntingz@usc.edu}{juntingz}, \href{mailto:lian455@usc.edu}{lian455}\}@usc.edu}, \texttt{\href{mailto:cckuo@sipi.usc.edu}{cckuo@sipi.usc.edu}}}}

\begin{document}
\ninept
\maketitle

\begin{abstract}
A domain adaptation method for urban scene segmentation is proposed in
this work. We develop a fully convolutional tri-branch network, where
two branches assign pseudo labels to images in the unlabeled target
domain while the third branch is trained with supervision based on images
in the pseudo-labeled target domain. The re-labeling and re-training
processes alternate. With this design, the tri-branch network learns
target-specific discriminative representations progressively and, as a
result, the cross-domain capability of the segmenter improves. We
evaluate the proposed network on large-scale domain adaptation
experiments using both synthetic (GTA) and real (Cityscapes) images. It
is shown that our solution achieves the state-of-the-art performance and
it outperforms previous methods by a significant margin. 
\end{abstract}
\section{Introduction}\label{sec:intro}

Semantic segmentation for urban scenes is an important yet challenging
task for a variety of vision-based applications, including autonomous
driving cars, smart surveillance systems, etc. With the success of
convolutional neural networks (CNNs), numerous successful
fully-supervised semantic segmentation solutions have been proposed in
recent years \cite{long2015fully, chen2016deeplab}. To achieve
satisfactory performance, these methods demand a sufficiently large
dataset with pixel-level labels for training. However, creating such
large datasets is prohibitively expensive as it requires human
annotators to accurately trace segment boundaries. Furthermore, it is
difficult to collect traffic scene images with sufficient variations in
terms of lighting conditions, weather, city and driving routes. 

To overcome the above-mentioned limitations, one can utilize the modern
urban scene simulators to automatically generate a large amount of synthetic images
with pixel-level labels. However, this introduces another
problem, \emph{i.e.} distributions mismatch between the source domain
(synthesized data) and the target domain (real data). Even if we
synthesize images with the state-of-the-art simulators
\cite{richter2016playing, ros2016synthia}, there still exists visible
appearance discrepancy between these two domains. The testing
performance in the target domain using the network trained solely by the
source domain images is severely degraded. The domain adaptation (DA)
technique is developed to bridge this gap. It is a special example of
transfer learning that leverages labeled data in the source domain to
learn a robust classifier for unlabeled data in the target domain.  DA
methods for object classification have several challenges such as shifts
in lighting and variations in object's appearance and pose. 
There are even more challenges in DA methods for semantic segmentation
because of variations in the scene layout, object scales and class
distributions in images. Many successful domain-alignment-based methods
work for DA-based classification but not for DA-based segmentation.
Since it is not clear what comprises data instances in a deep segmenter
\cite{Zhang_2017_ICCV}, DA-based segmentation is still far from its
maturity. 

In this work, we propose a novel fully convolutional tri-branch network
(FCTN) to solve the DA-based segmentation problem.  In the FCTN, two
labeling branches are used to generate pseudo segmentation ground-truth
for unlabeled target samples while the third branch learns from these
pseudo-labeled target samples. An alternating re-labeling and
re-training mechanism is designed to improve the DA performance in a
curriculum learning fashion. We evaluate the proposed method using
large-scale synthesized-to-real urban scene datasets and demonstrate
substantial improvement over the baseline network and other benchmarking
methods. 

\begin{figure*}[!t]
\centering
\includegraphics[width=0.7\linewidth]{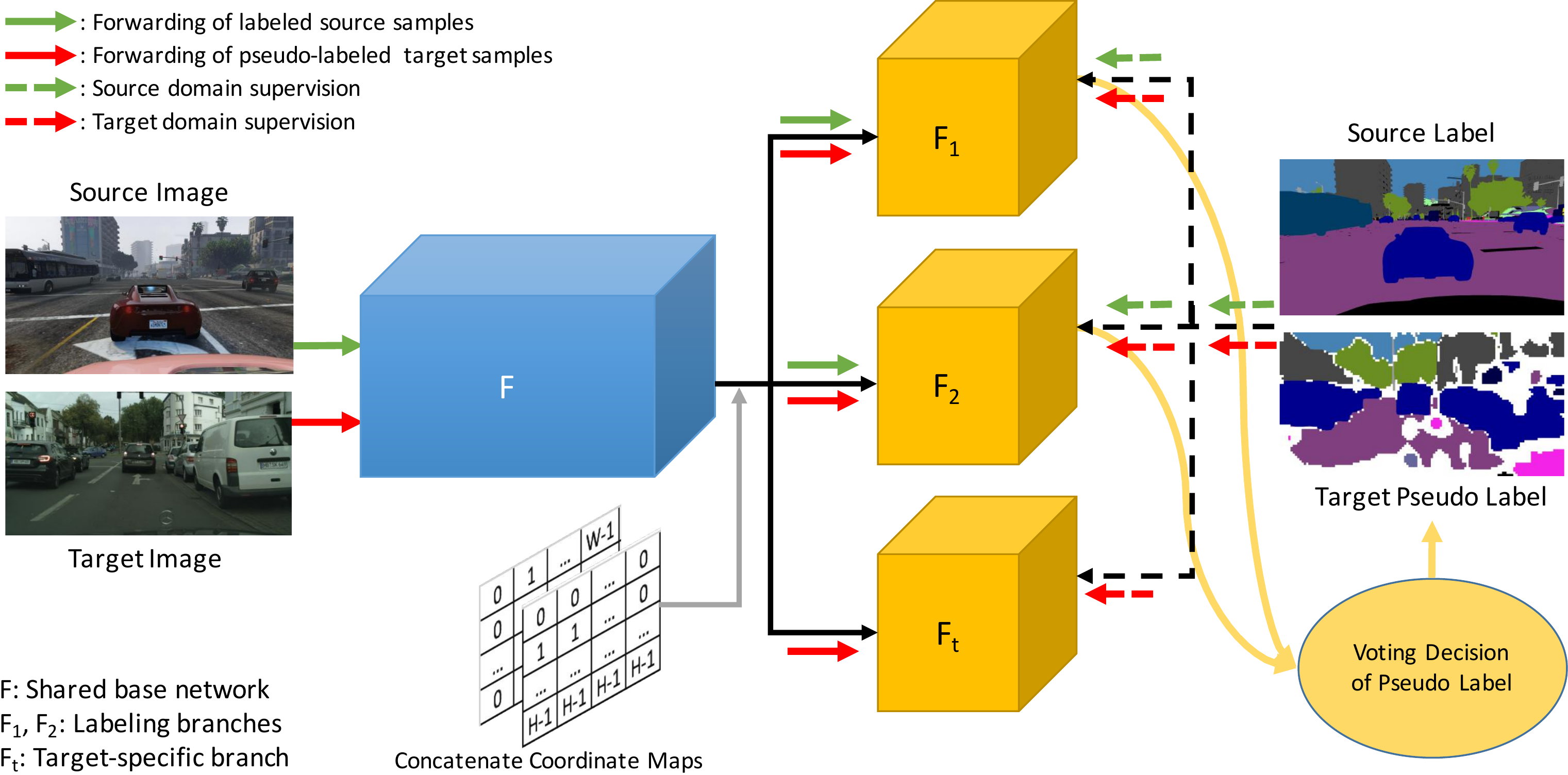}
\vspace{0.5em}
\caption{An overview of the proposed fully convolutional tri-branch
network (FCTN). It has one shared base network denoted by $F$ followed
by three branches of the same architecture denoted by $F_1$, $F_2$ and
$F_t$. Branches $F_1$ and $F_2$ assign pseudo labels to images in the
unlabeled target domain, while branch $F_t$ is trained with supervision
from images in the pseudo-labeled target domain.}
\label{fig:overview}
\vspace{-1em}
\end{figure*}

\section{Related Work}\label{sec:related_work}

The current literatures on visual domain adaptation mainly focus on
image classification \cite{Gabriela2017DABook}. Being inspired by
shallow DA methods, one common intuition of deep DA methods is that
adaptation can be achieved by matching the distribution of features in
different domains. Most deep DA methods follow a siamese architecture
with two streams, representing the source and target models. They aim to
obtain domain-invariant features by minimizing the divergence of
features in the two domains and a classification loss
\cite{long2015learning, sun2016deep, tzeng2015simultaneous,
tzeng2017adversarial}, where the classification loss is evaluated in the
source domain with labeled data only. However, these methods assume the
existence of a universal classifier that can perform well on samples
drawn from whichever domain. This assumption tends to fail since the
class correspondence constraint is rarely imposed in the domain
alignment process. Without such an assumption, feature distribution
matching may not lead to classification improvement in the target
domain. The ATDA method proposed in \cite{saito2017asymmetric} avoids
this assumption by employing the asymmetric tri-training. It can assign
pseudo labels to unlabeled target samples progressively and learn from
them using a curriculum learning paradigm. 
This paradigm has been proven effective in the weakly-supervised learning tasks \cite{li2017multiple} as well.

Previous work on segmentation-based DA is much less. Hoffman \emph{et.
al} \cite{hoffman2016fcns} consider each spatial unit in an activation
map of a fully convolutional network (FCN) as an instance, and extend
the idea in \cite{tzeng2015simultaneous} to achieve two objectives: 1)
minimizing the global domain distance between two domains using a fully
convolutional adversarial training and 2) enhancing category-wise
adaptation capability via multiple instance learning. The adversarial
training aims to align intermediate features from two domains.  It
implies the existence of a single good mapping from the domain-invariant
feature space to the correct segmentation mask.  To avoid this
condition, Zhang \emph{et. al} \cite{Zhang_2017_ICCV} proposed to
predict the class distribution over the entire image and some
representative super pixels in the target domain first. Then, they use
the predicted distribution to regularize network training. In this work,
we avoid the single good mapping assumption and rely on the remarkable
success of the ATDA method \cite{saito2017asymmetric}. In particular,
we develop a curriculum-style method that improves the cross-domain
generalization ability for better performance in DA-based segmentation. 

\section{Proposed Domain Adaptation Network}\label{sec:approach}

The proposed fully convolutional tri-branch network (FCTN) model for
cross-domain semantic segmentation is detailed in this section.  The
labeled source domain training set is denoted by $\mathcal{S} =
\{(x_i^s, y_i^s)\}_{i=1}^{n_s}$ while the unlabeled target domain
training set is denoted by $\mathcal{T} = \{x_i^t\}_{i=1}^{n_t}$, where
$x$ is an image, $y$ is the ground truth segmentation mask and $n_s$ and
$n_t$ are the sizes of training sets of two domains, respectively. 

\subsection{Fully Convolutional Tri-branch Network Architecture}\label{ssec:architecture}

An overview of the proposed FCTN architecture is illustrated in Fig.
\ref{fig:overview}. It is a fully convolutional network that consists of
a shared base network ($F$) followed by three branch networks ($F_1$,
$F_2$ and $F_t$). Branches $F_1$ and $F_2$ are labeling branches. They
accept deep features extracted by the shared base net, $F$, as the input
and predict the semantic label of each pixel in the input image.
Although the architecture of the three branches are the same, their
roles and functions are not identical.  $F_1$ and $F_2$ generate pseudo
labels for the target images based on prediction. $F_1$ and $F_2$ learn
from both labeled source images and pseudo-labeled target images. In
contrast, $F_t$ is a target-specific branch that learns from
pseudo-labeled target images only. 

We use the DeepLab-LargeFOV (also known as the DeepLab v1)
\cite{chen2015semantic} as the reference model due to its simplicity and
superior performance in the semantic segmentation task. The
DeepLab-LargeFOV is a re-purposed VGG-16 \cite{simonyan2014very} network
with dilated convolutional kernels. The shared base network $F$ contains
13 convolutional layers while the three branche networks are formed by
three convolutional layers that are converted from fully connected
layers in the original VGG-16 network. Although the DeepLab-LargeFOV is
adopted here, any effective FCN-based semantic segmentation framework
can be used in the proposed FCTN architecture as well. 

\subsection{Encoding Explicit Spatial Information}

Being inspired by PFN \cite{liang2015proposal}, we attach the pixel
coordinates as the additional feature map to the last layer of $F$.  The
intuition is that the urban traffic scene images have structured layout
and certain classes usually appear in a similar location in images.
However, a CNN is translation-invariant by nature. That is, it makes
prediction based on patch-based feature regardless of the patch
location in the original image. Assume that the last layer in $F$ has a
feature map of size $H \times W \times D$, where $H$, $W$ and $D$ are
the height, width and depth of the feature map, respectively. We
generate two spatial coordinate maps $\mat{X}$ and $\mat{Y}$ of size $H
\times W$, where values of $\mat{X}(p_x,p_y)$ and $\mat{Y}(p_x,p_y)$ are
set to be $p_x/W$ and $p_y/H$ for pixel $p$ at location $(p_x,p_y)$, respectively. We
concatenate spatial coordinate maps $\mat{X}$ and $\mat{Y}$ to the
original feature maps along the depth dimension. Thus, the output
feature maps are of dimension $H \times W \times (D+2)$. By
incorporating the spatial coordinate maps, the FCTN can learn more
location-aware representations. 

\subsection{Assigning Pseudo Labels to Target Images}\label{ssec:labeling}

\begin{figure}[htb]	
\centering
\includegraphics[width=.32\linewidth]{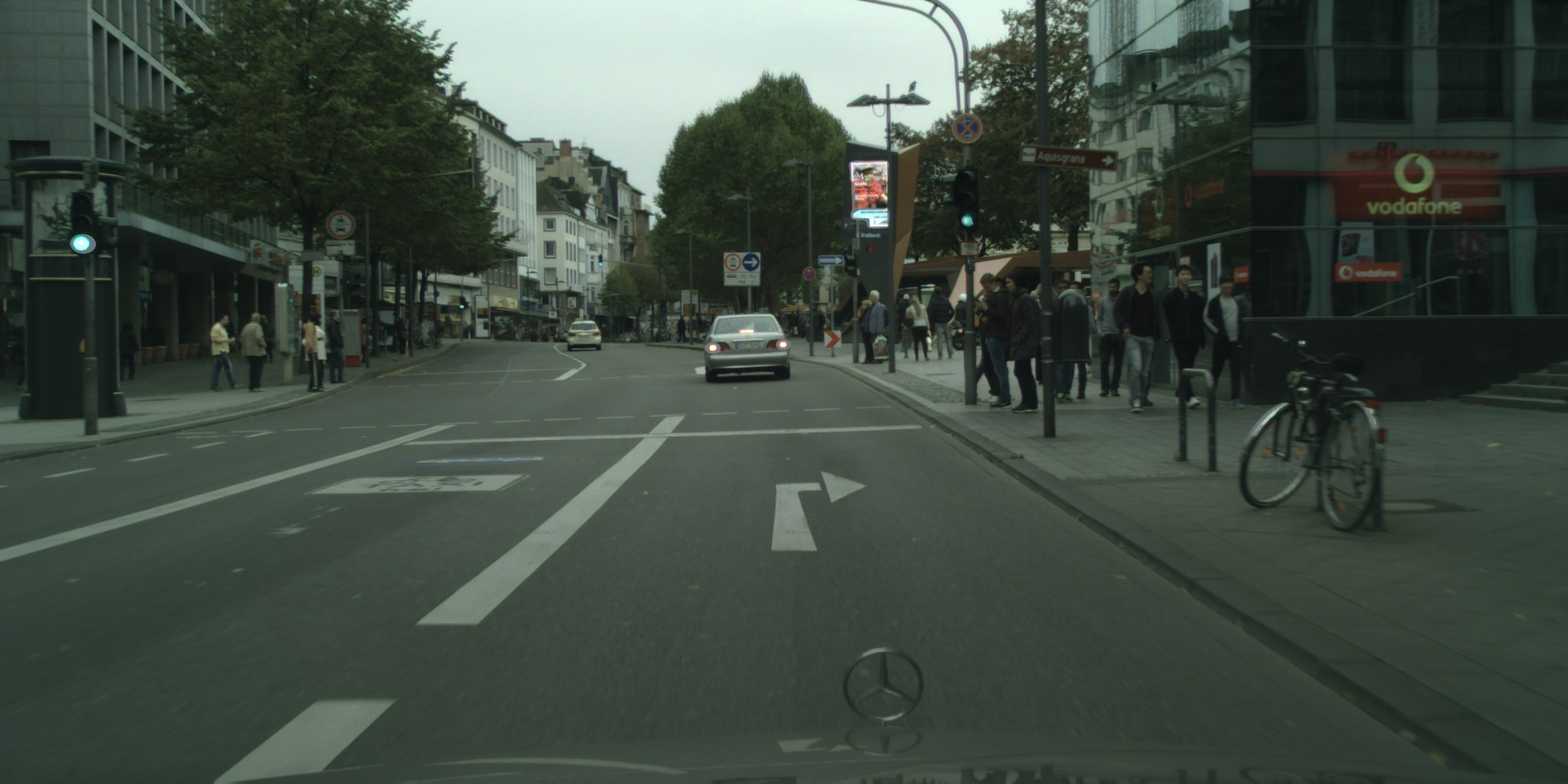}
\includegraphics[width=.32\linewidth]{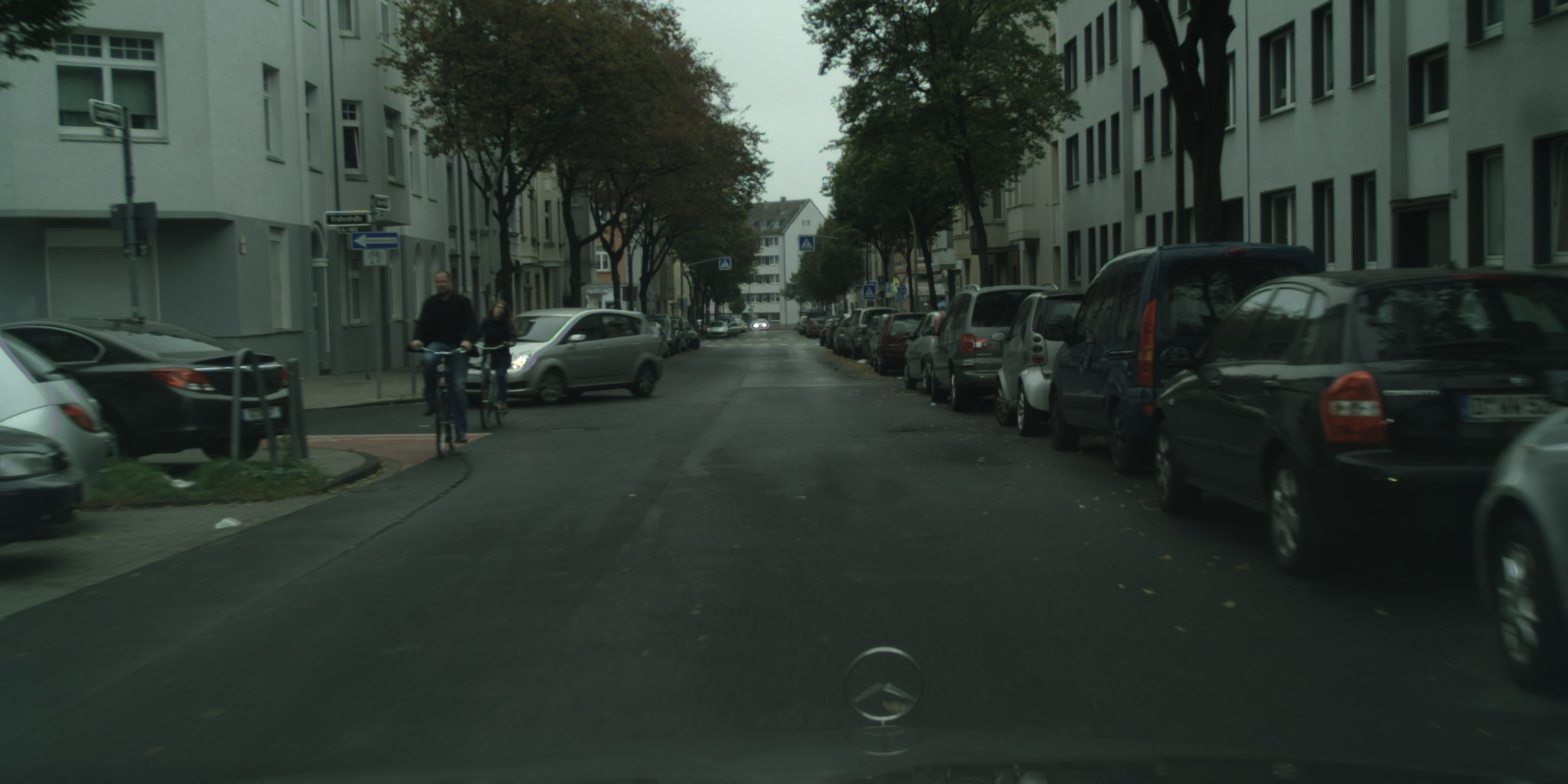}
\includegraphics[width=.32\linewidth]{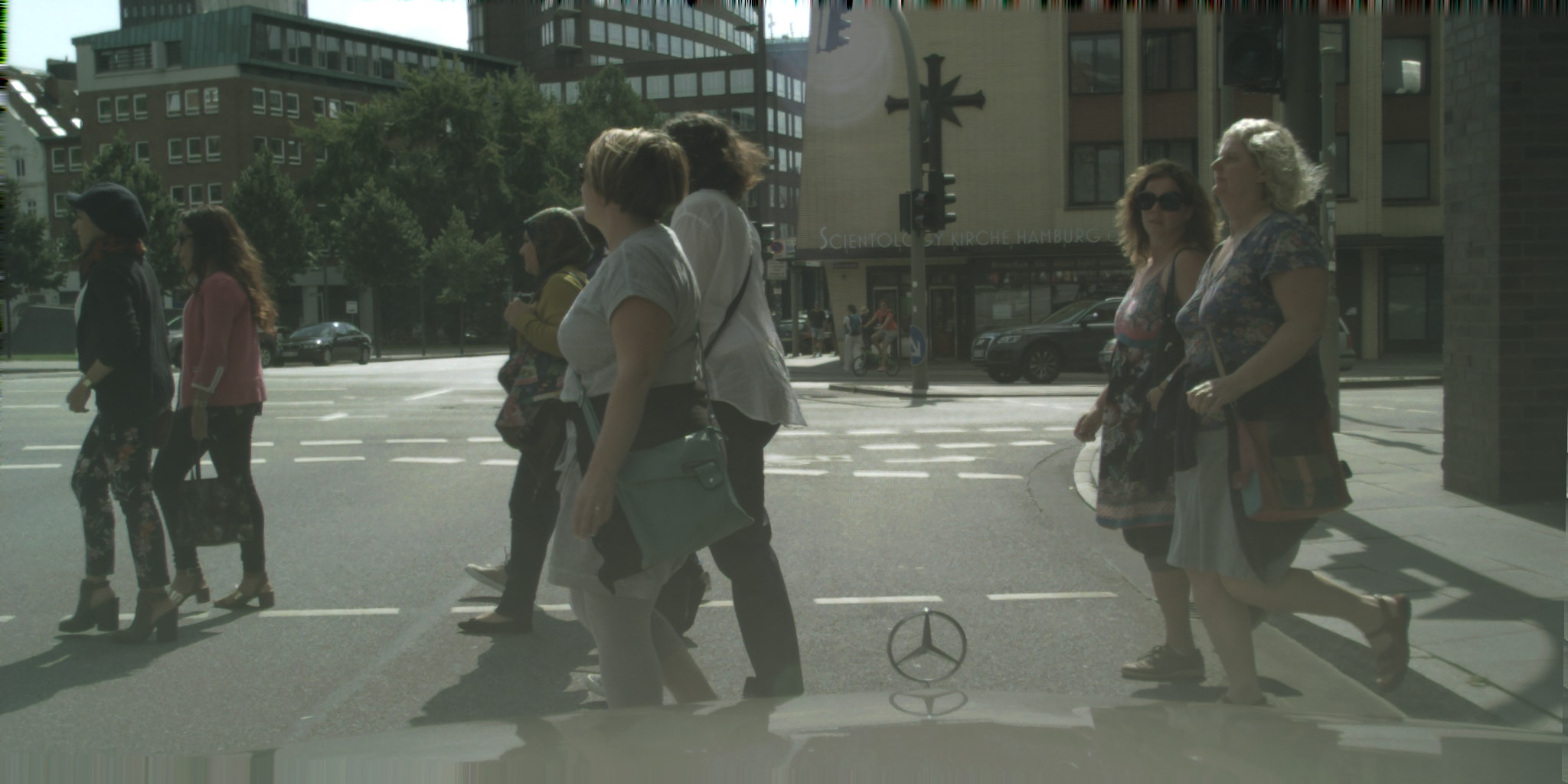}
	
\includegraphics[width=.32\linewidth]{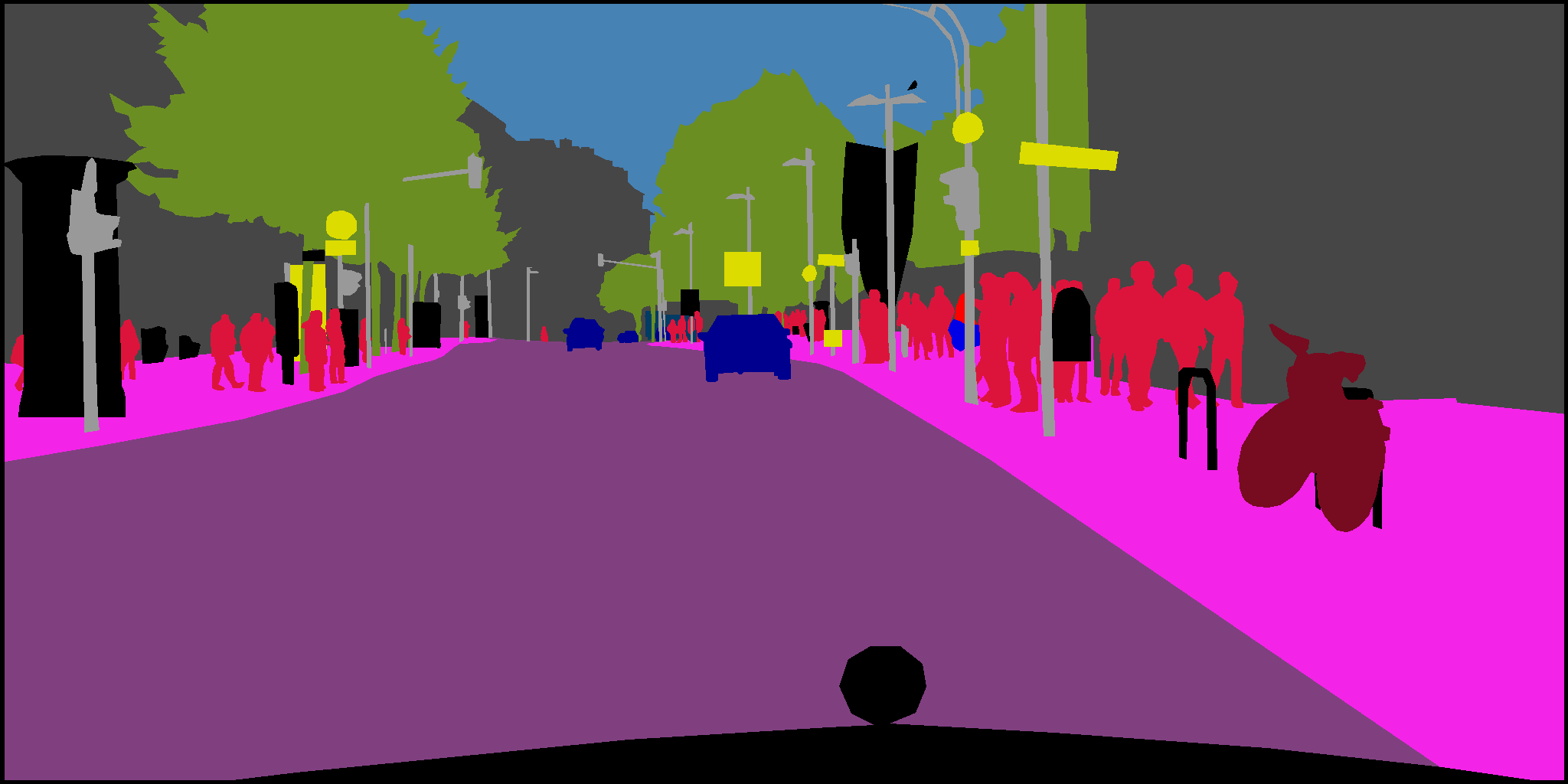}
\includegraphics[width=.32\linewidth]{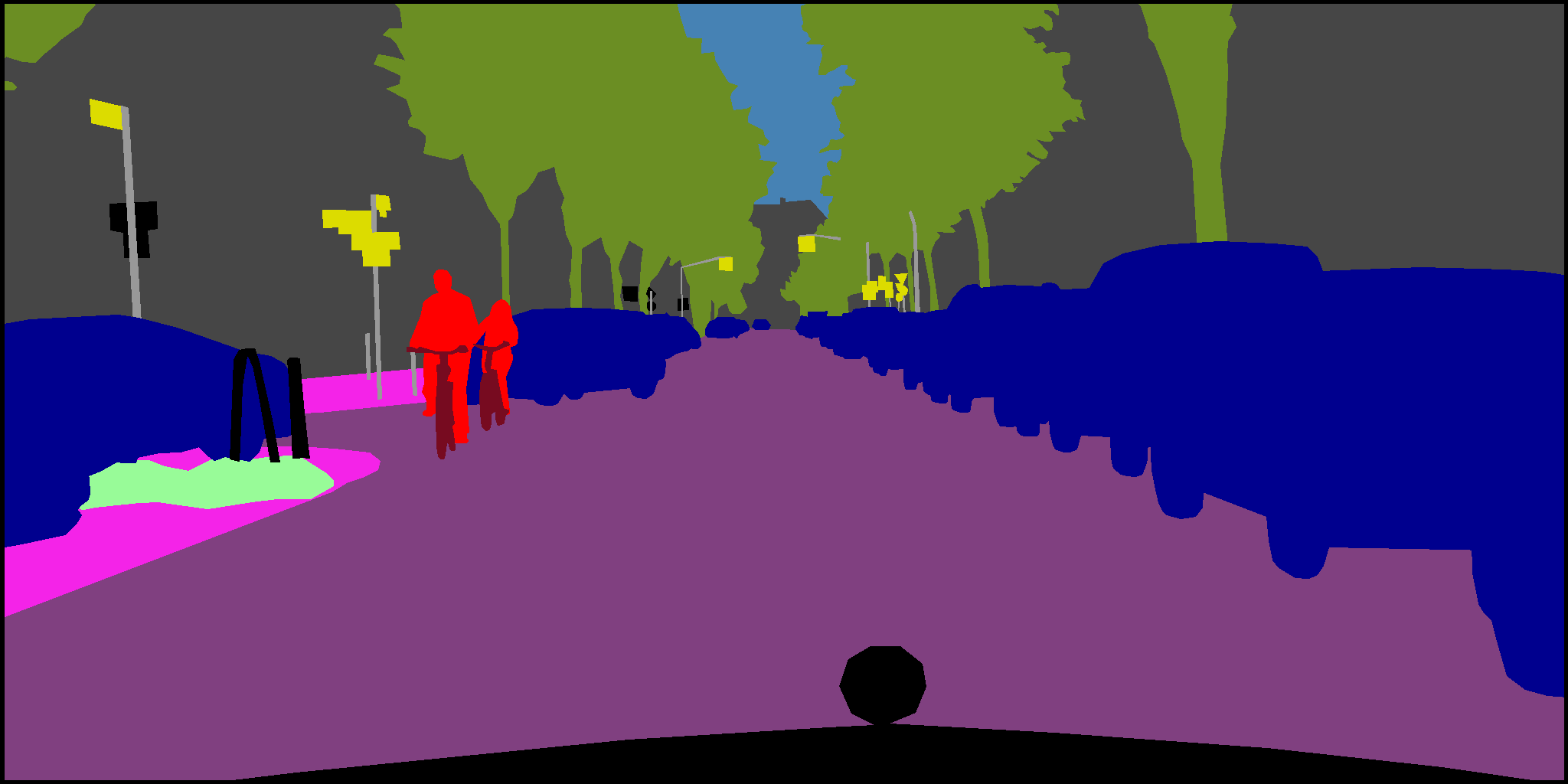}
\includegraphics[width=.32\linewidth]{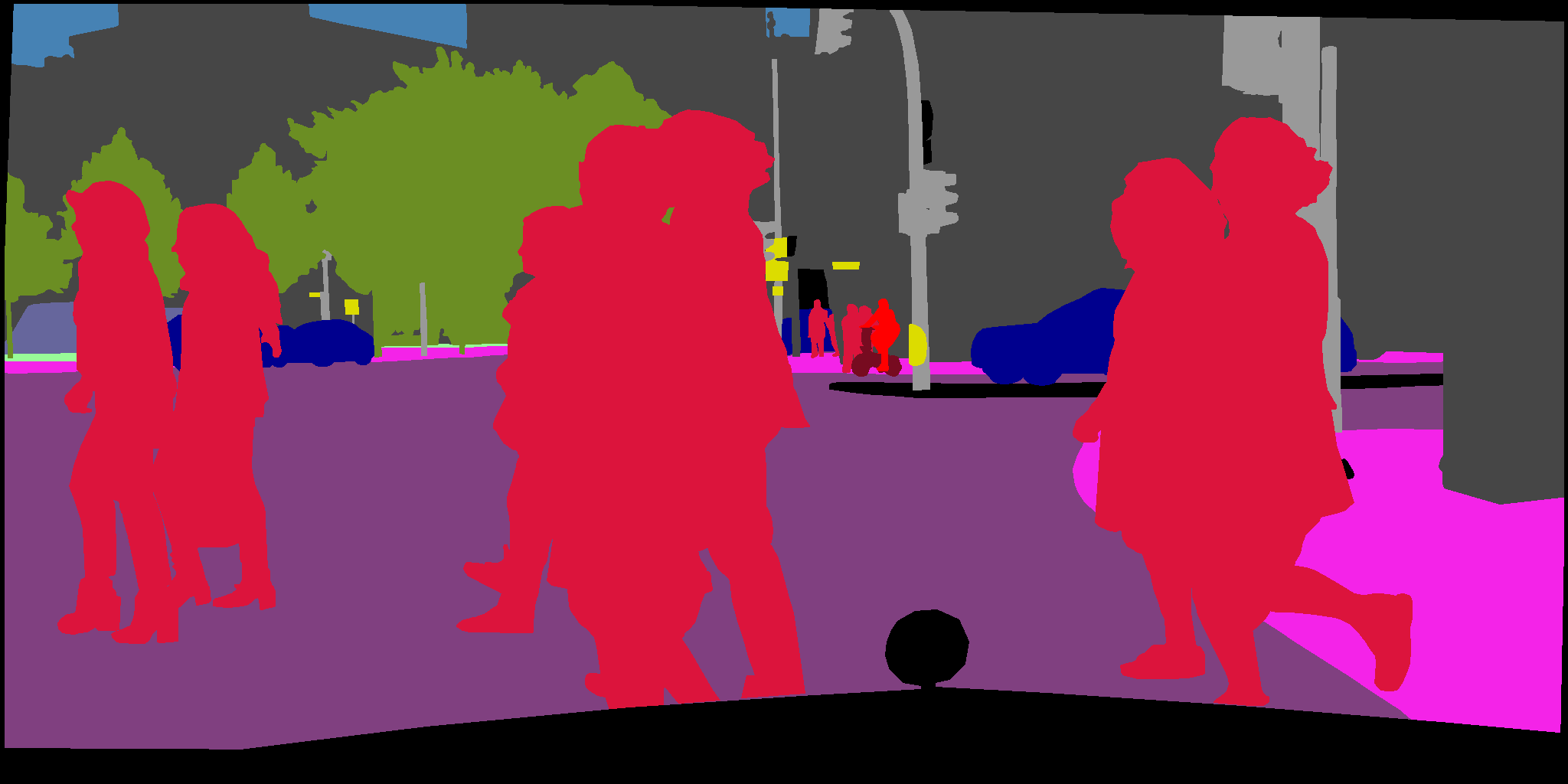}
	
\includegraphics[width=.32\linewidth]{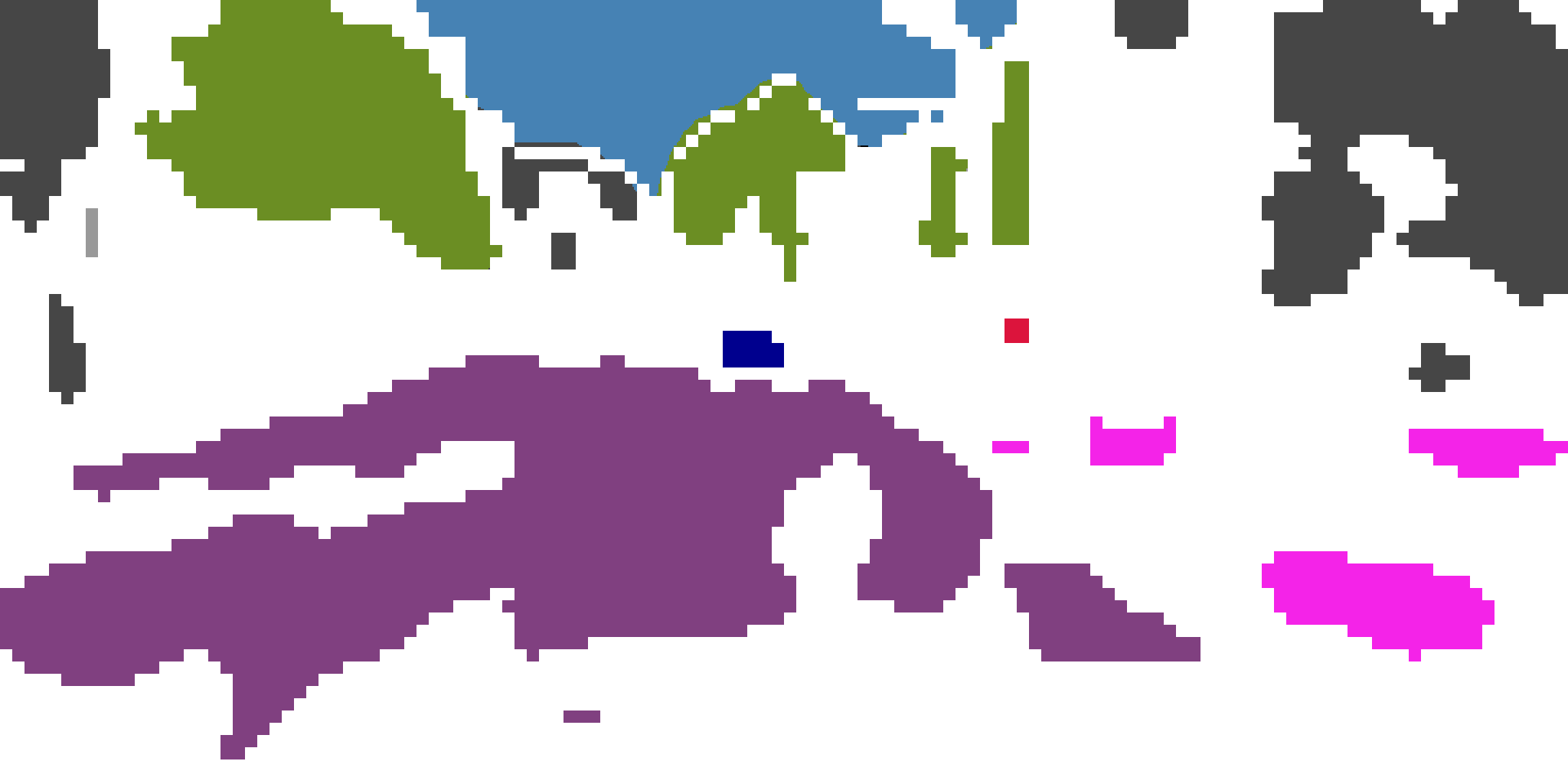}
\includegraphics[width=.32\linewidth]{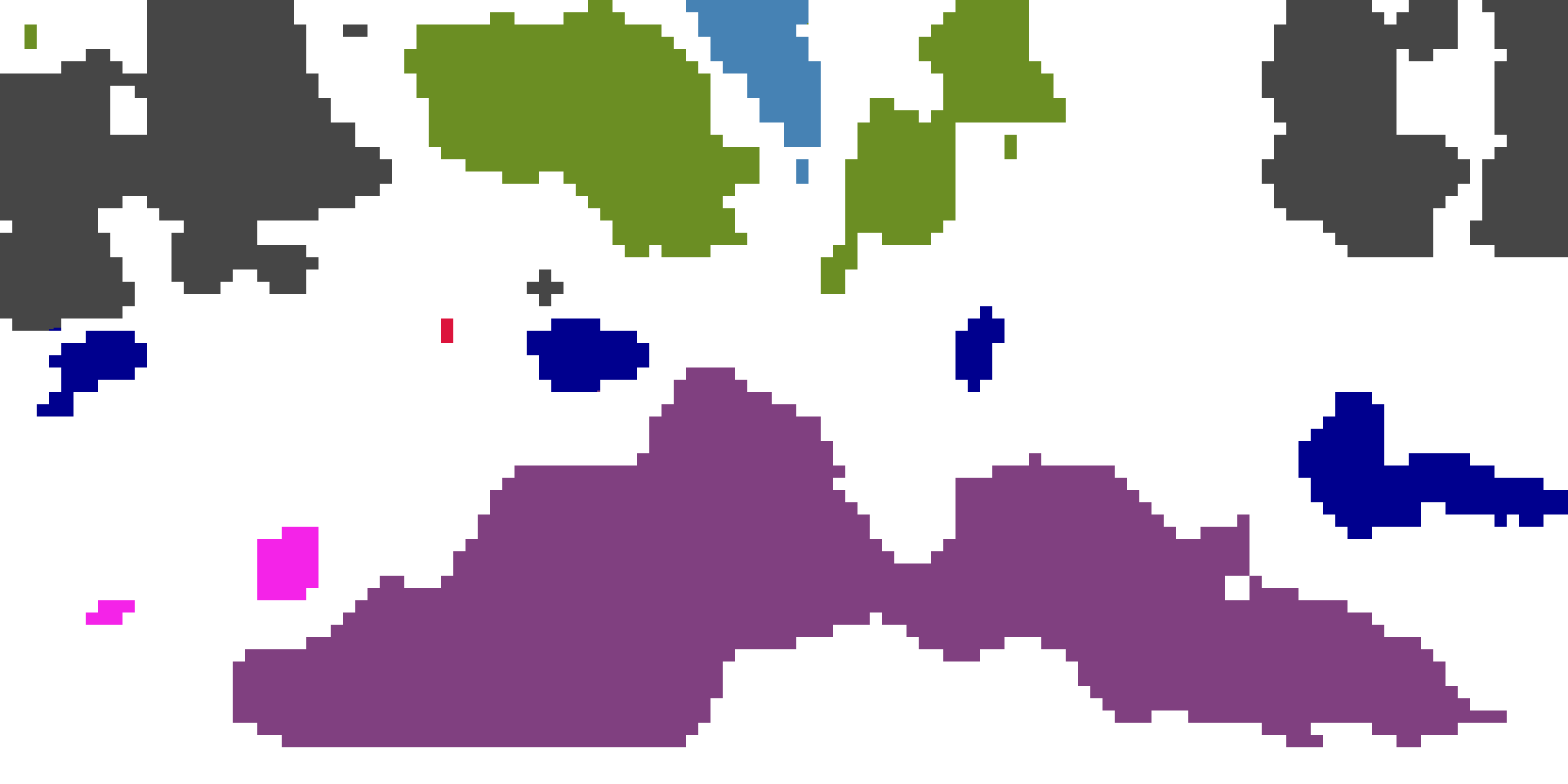}
\includegraphics[width=.32\linewidth]{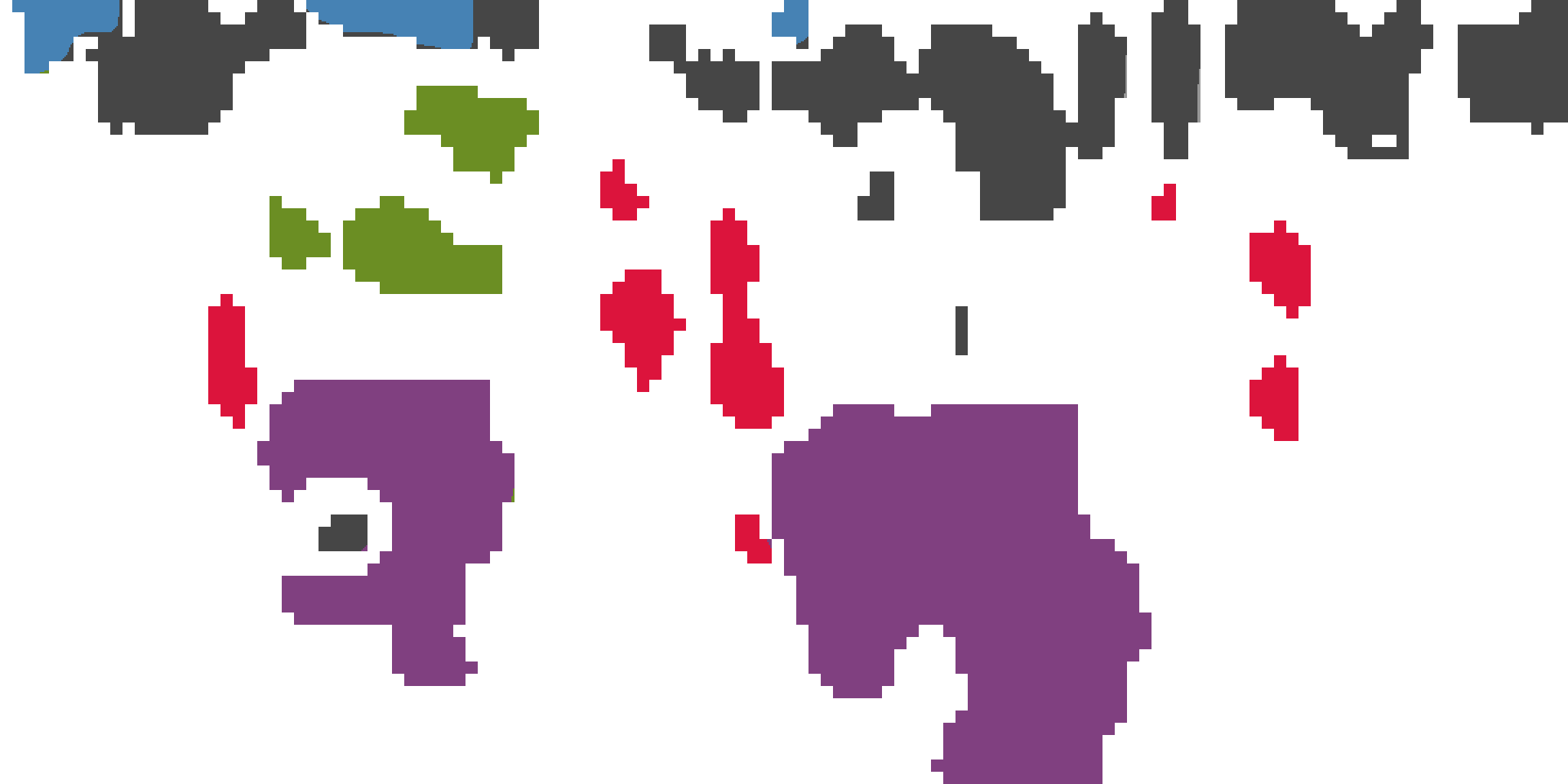}
	
\includegraphics[width=.32\linewidth]{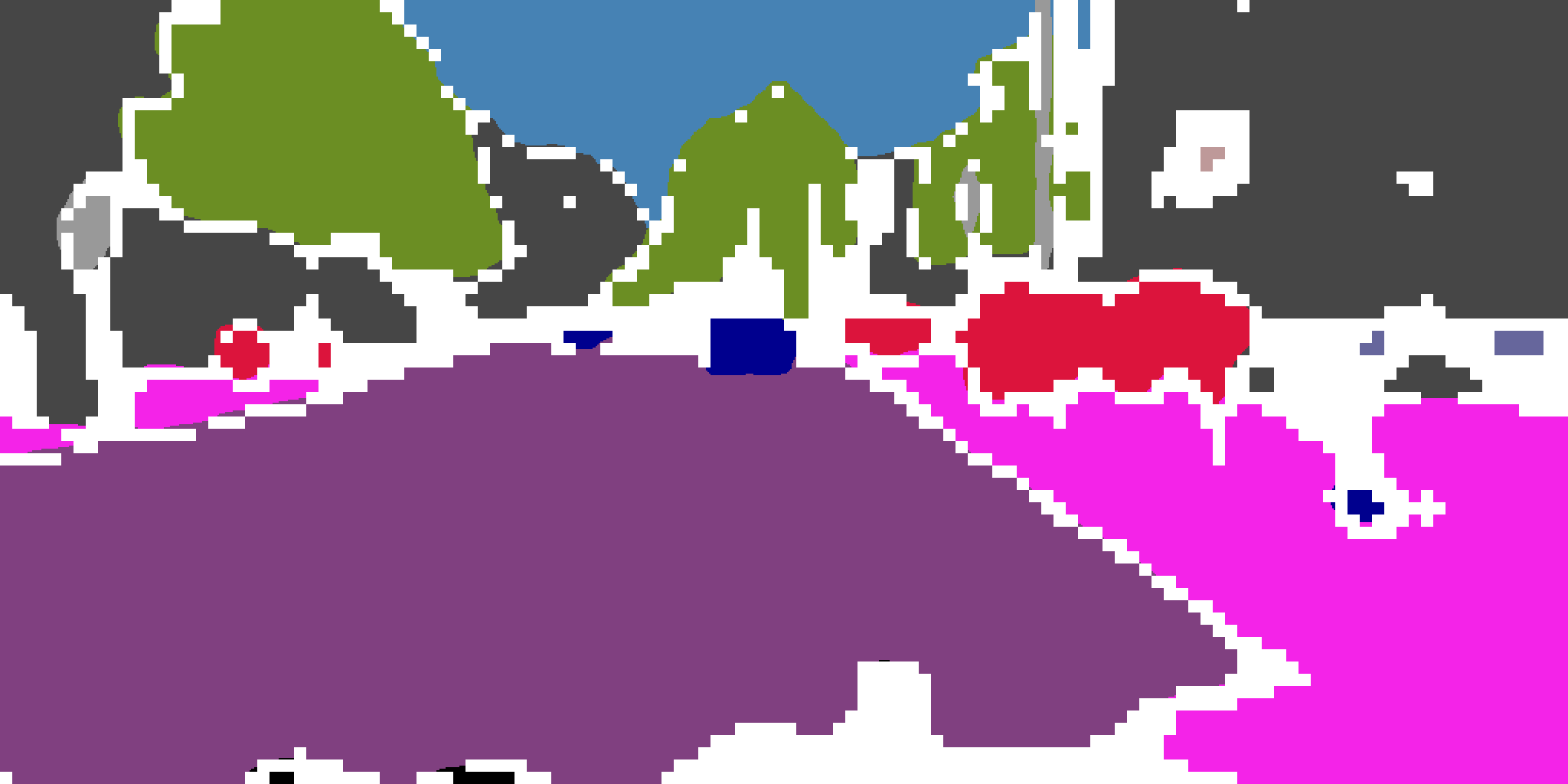}	
\includegraphics[width=.32\linewidth]{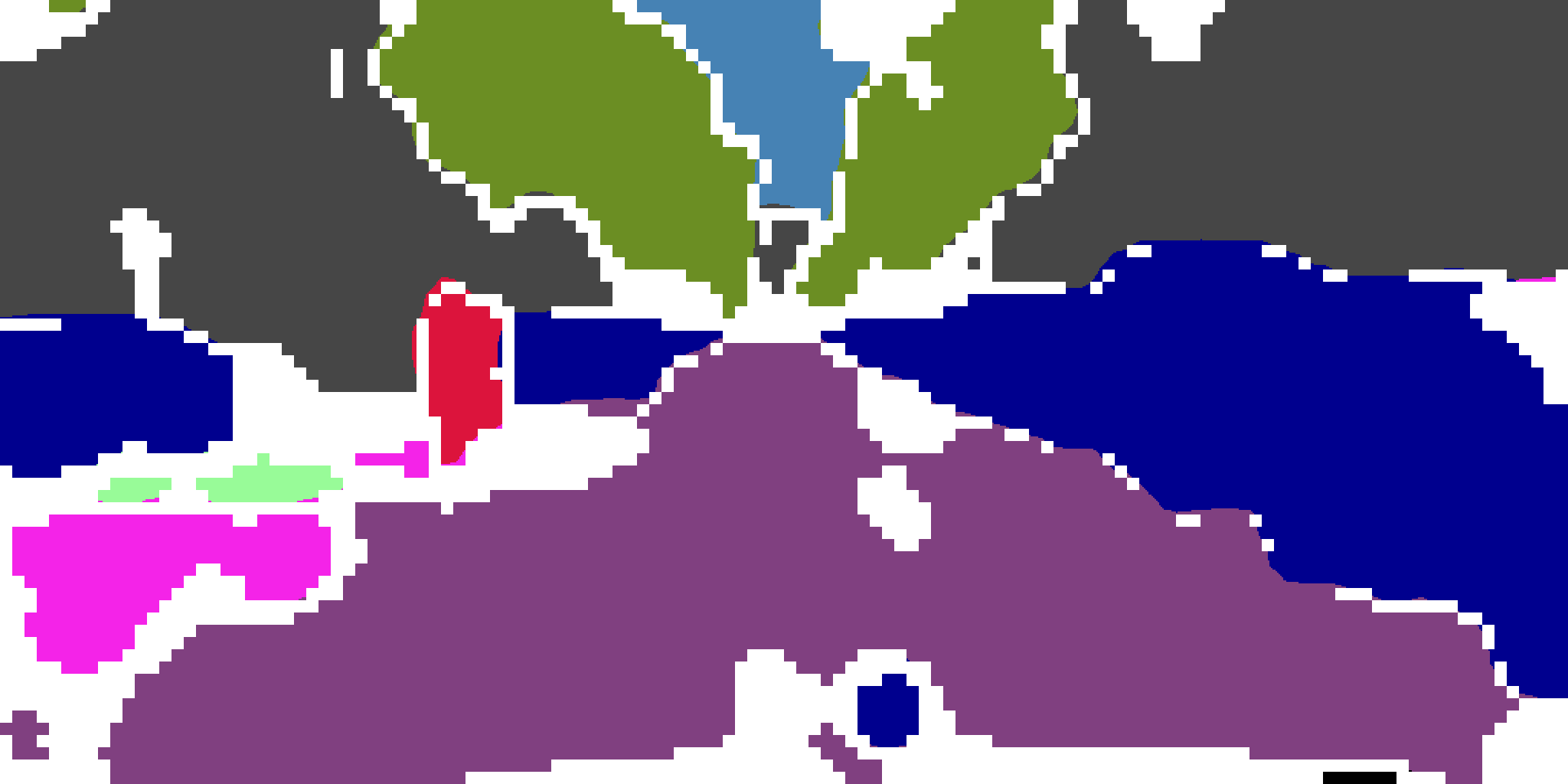}	
\includegraphics[width=.32\linewidth]{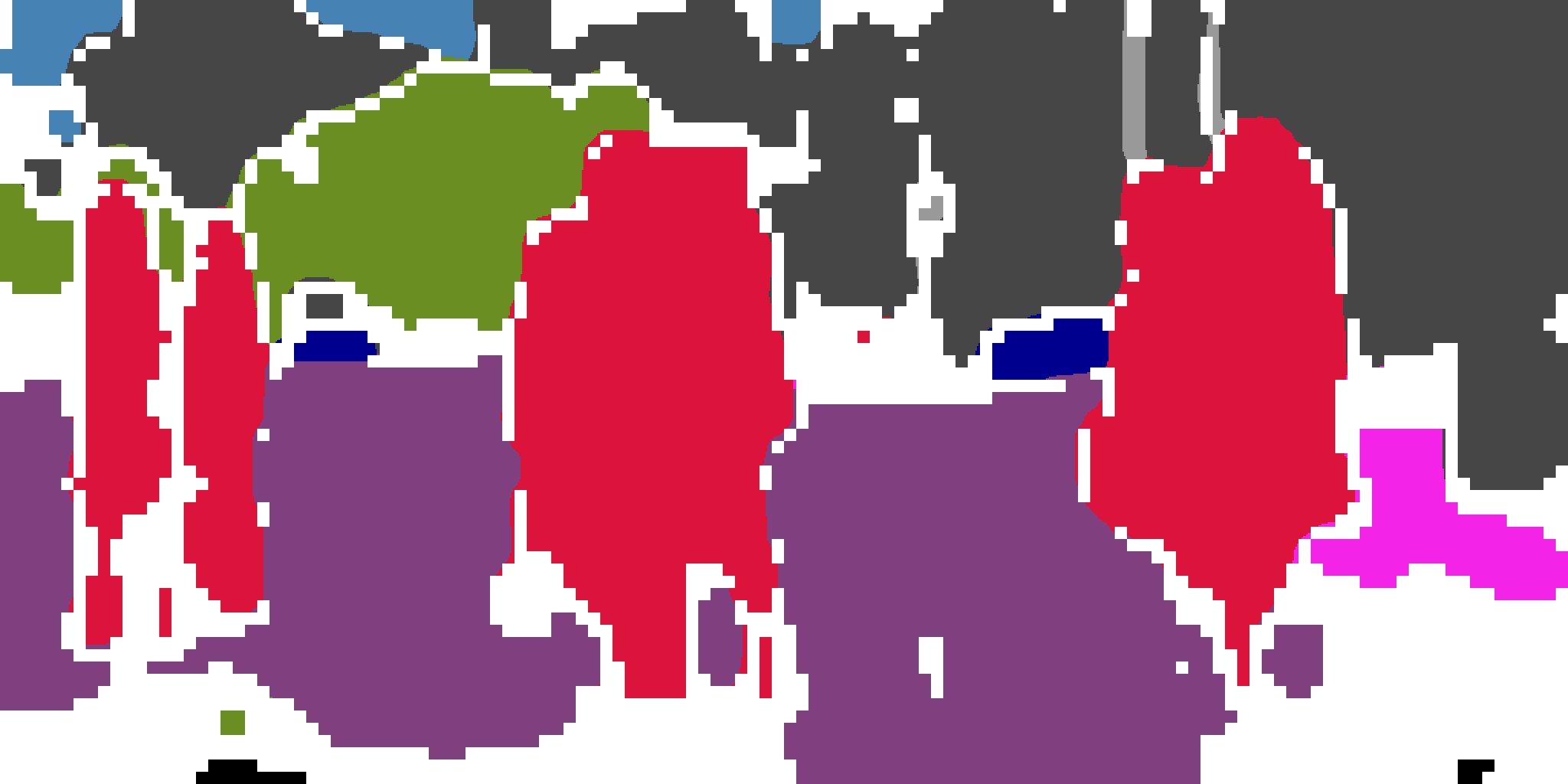}
\vspace{0.5em}
\caption{Illustration of pseudo labels used in the 2-round curriculum
learning in the GTA-to-Cityscapes DA experiments. The first row shows
the input images.  The second row shows the ground truth segmentation
masks. The third and fourth row shows the pseudo labels used in the
first and second round of curriculum learning, respectively. Note in the
visualization of pseudo labels, white pixels indicate the unlabeled pixels. Best
viewed in color.} \label{fig:pseudo_label}
\vspace{-1em}
\end{figure}

Being inspired by the ATDA method \cite{saito2017asymmetric}, we
generate pseudo labels by feeding images in the target domain training
set to the FCTN and collect predictions from both labeling branches.
For each input image, we assign the pseudo-label to a pixel if the
following two conditions are satisfied: 1) the classifiers associated
with labeling branches, $F_1$ and $F_2$, agree in their predicted labels
on this pixel; 2) the higher confidence score of these two predictions
exceeds a certain threshold. In practice, the confidence threshold is
set very high (say, 0.95 in our implementation) because the use of many
inaccurate pseudo labels tends to mislead the subsequent network
training. In this way, high-quality pseudo labels for target images are
used to guide the network to learn target-specific discriminative
features. The pseudo-labeled target domain training set is denoted
by $\mathcal{T}_l=\{(x_i^t, \hat{y}_i^t)\}_{i=1}^{n_t}$, where
$\hat{y}$ is the partially pseudo-labeled segmentation mask.  Some
sample pseudo-labeled segmentation masks are shown in Fig.
\ref{fig:pseudo_label}. In the subsequent training, the not-yet-labeled
pixels are simply ignored in the loss computation. 

\subsection{Loss Function}\label{ssec:loss}

\textbf{Weight-Contrained Loss.} As suggested in the standard
tri-training algorithm \cite{zhou2005tri}, the three classifiers in
$F_1$, $F_2$ and $F_t$ must be diverse. Otherwise, the training
degenerates to self-training. In our case, one crucial requirement to
obtain high-quality pseudo-labels from two labeling branches $F_1$ and
$F_2$ is that they should have different views on one sample and make decisions
on their own. 

Unlike the case in the co-training algorithm \cite{blum1998combining},
where one can explicitly partition features into different sufficient
and redundant views, it is not clear how to partition deep features
effectively in our case. Here, we enforce divergence of the weights of
the convolutional layers of two labeling branches by minimizing their
cosine similarity.  Then, we have the following filter weight-constrained
loss term:
\begin{equation}
L_w = \frac{\inner{\vec{w_1}}{\vec{w_2}}}{\norm{\vec{w_1}}\norm{\vec{w_2}}}
\end{equation}
where $\vec{w_1}$ and $\vec{w_2}$ are obtained by the flattening and
concatenating the weights of convolutional filters in convolutional
layers of $F_1$ and $F_2$, respectively. 

\textbf{Weighted Pixel-wise Cross-entropy Loss.} In the curriculum
learning stage, we take a minibatch of samples with one half from
$\mathcal{S}$ and the other half from $\mathcal{T}_l$ at each step. We
calculate the segmentation losses separately for each half of samples.
For the source domain images samples, we use the vanilla pixel-wise
softmax cross-entropy loss, denoted by $L_{\mathcal{S}}$, as the
segmentation loss function. 

Furthermore, as mentioned in Sec.
\ref{ssec:labeling}, we assign pseudo labels to target domain pixels
based on predictions of two labeling branches.  This mechanism tends to
assign pseudo labels to the prevalent and easy-to-predict classes, such
as the road, building, etc., especially in the early stage (this can be seen in Fig. \ref{fig:pseudo_label}). Thus, the
pseudo labels can be highly imbalanced in classes. If we treat all
classes equally, the gradients from challenging and relatively rare
classes will be insignificant and the training will be biased toward
prevalent classes. To remedy this, we use a weighted cross-entropy loss
for target domain samples, denoted by $L_{\mathcal{T}_l}$. We calculate
weights using the median frequency balancing scheme
\cite{eigen2015predicting}, where the weight assigned to class $c$ in
the loss function becomes
\begin{equation}
\alpha_c = \frac{median\_freq}{freq(c)},
\end{equation}
where $freq(c)$ is the number of pixels of class $c$ divided by the
total number of pixels in the source domain images whenever $c$ is
present, and $median\_freq$ is the median of these frequencies
$\{freq(c)\}_{c=1}^{C}$, and where $C$ is the total number of classes.
This scheme works well under the assumption that the global class
distributions of the source domain and the target domain are similar. 

\textbf{Total Loss Function.} There are two stages in our training
procedure. We first pre-train the entire network using minibatches from
$\mathcal{S}$ so as to minimize the following objective function:
\begin{equation}\label{eq:pretrain}
L = \alpha L_w + L_{\mathcal{S}}
\end{equation}
Once the curriculum learning starts, the overall objective function 
becomes
\begin{equation}\label{eq:curriculum}
L = \alpha L_w + L_{\mathcal{S}} + \beta L_{\mathcal{T}_l}
\end{equation} 
where $L_{\mathcal{S}}$ is evaluated on $\mathcal{S}$ and averaged over
predictions of $F_1$ and $F_2$ branches, $L_{\mathcal{T}_l}$ is
evaluated on $\mathcal{T}_l$ and averaged over predictions of all three
top branches, and $\alpha$ and $\beta$ are hyper-parameters determined
by the validation split. 

\begin{table*}[htb]
	\centering
	\setlength\tabcolsep{3pt}
	\begin{tabular}{|c|c|c|c|c|c|c|c|c|c|c|c|c|c|c|c|c|c|c|c|c|}
		\hline
		\multirow{2}[4]{*}{Model} & \multicolumn{19}{c|}{per-class IoU}                                                                                                                   & \multirow{2}[4]{*}{mIoU} \\
		\cline{2-20}          & \begin{sideways}road\end{sideways} & \begin{sideways}sidewlk\end{sideways} & \begin{sideways}bldg.\end{sideways} & \begin{sideways}wall\end{sideways} & \begin{sideways}fence\end{sideways} & \begin{sideways}pole\end{sideways} & \begin{sideways}t. light\end{sideways} & \begin{sideways}t. sign\end{sideways} & \begin{sideways}veg.\end{sideways} & \begin{sideways}terr.\end{sideways} & \begin{sideways}sky\end{sideways} & \begin{sideways}person\end{sideways} & \begin{sideways}rider\end{sideways} & \begin{sideways}car\end{sideways} & \begin{sideways}truck\end{sideways} & \begin{sideways}bus\end{sideways} & \begin{sideways}train\end{sideways} & \begin{sideways}mbike\end{sideways} & \begin{sideways}bike\end{sideways} &  \\
		\hline
		No Adapt & 31.9  & 18.9  & 47.7  & 7.4   & 3.1   & 16.0  & 10.4  & 1.0   & 76.5  & 13.0  & 58.9  & 36.0  & 1.0   & 67.1  & 9.5   & 3.7   & 0.0   & 0.0   & 0.0   & 21.1 \\
		
		FCN \cite{hoffman2016fcns} & 70.4  & \textbf{32.4} & 62.1  & 14.9  & 5.4   & 10.9  & 14.2  & 2.7   & 79.2  & 21.3  & 64.6  & 44.1  & 4.2   & 70.4  & 8.0   & 7.3   & 0.0   & 3.5   & 0.0   & 27.1 \\
		\hline
		No Adapt & 18.1  & 6.8   & 64.1  & 7.3   & 8.7   & 21.0  & 14.9  & 16.8  & 45.9  & 2.4   & 64.4  & 41.6  & \textbf{17.5} & 55.3  & 8.4   & 5.0   & \textbf{6.9} & 4.3   & 13.8  & 22.3 \\
		CDA \cite{Zhang_2017_ICCV} & 26.4  & 22.0  & 74.7  & 6.0   & \textbf{11.9} & 8.4   & 16.3  & 11.1  & 75.7  & 13.3  & \textbf{66.5} & 38.0  & 9.3   & 55.2  & \textbf{18.8} & \textbf{18.9} & 0.0   & \textbf{16.8} & \textbf{14.6} & 27.8 \\
		\hline
		No Adapt & 59.7  & 24.8  & 66.8  & 12.8  & 7.9   & 11.9  & 14.2  & 4.2   & 78.7  & 22.3  & 65.2  & 44.1  & 2.0   & 67.8  & 9.6   & 2.4   & 0.6   & 2.2   & 0.0   & 26.2 \\
		
		Round 1 & 66.9  & 25.6  & 74.7  & 17.5  & 10.3  & 17.1  & 18.4  & 8.0   & 79.7  & 34.8  & 59.7  & \textbf{46.7} & 0.0   & 77.1  & 10.0  & 1.8   & 0.0   & 0.0   & 0.0   & 28.9 \\
		
		Round 2 & \textbf{72.2} & 28.4  & \textbf{74.9} & \textbf{18.3} & 10.8  & \textbf{24.0} & \textbf{25.3} & \textbf{17.9} & \textbf{80.1} & \textbf{36.7} & 61.1  & 44.7  & 0.0   & \textbf{74.5} & 8.9   & 1.5   & 0.0   & 0.0   & 0.0   & \textbf{30.5} \\
		\hline
	\end{tabular}%
\caption{Adaptation from GTA to Cityscapes. All numbers are measured in \%. The last three
rows show our results before adaptation, after one and two rounds of curriculum learning using 
the proposed FCTN, respectively.}\label{tab:GTA}%
\end{table*}%

\begin{figure*}[tb!h]	
	\centering
	\includegraphics[width=.24\linewidth]{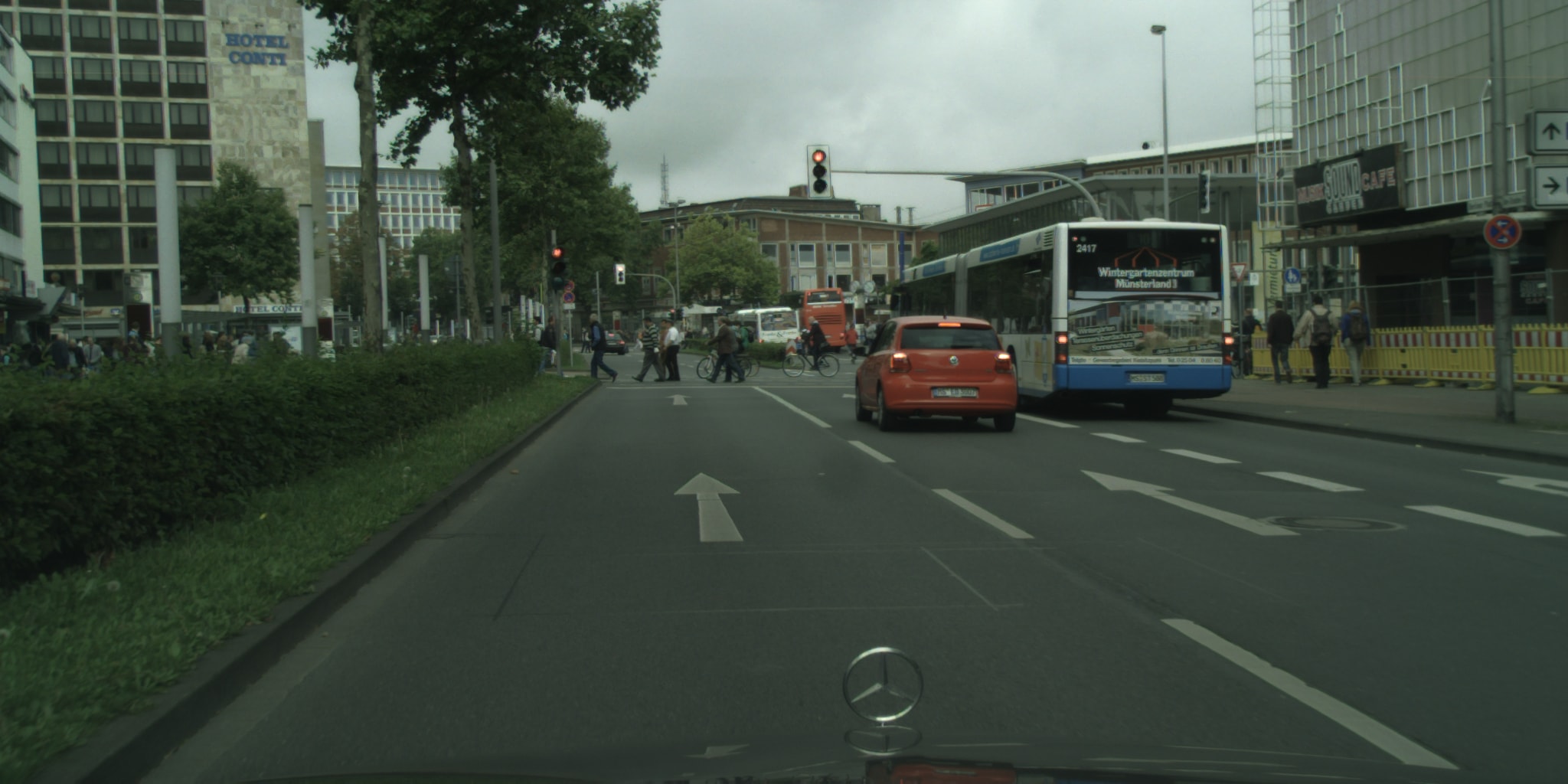}
	\includegraphics[width=.24\linewidth]{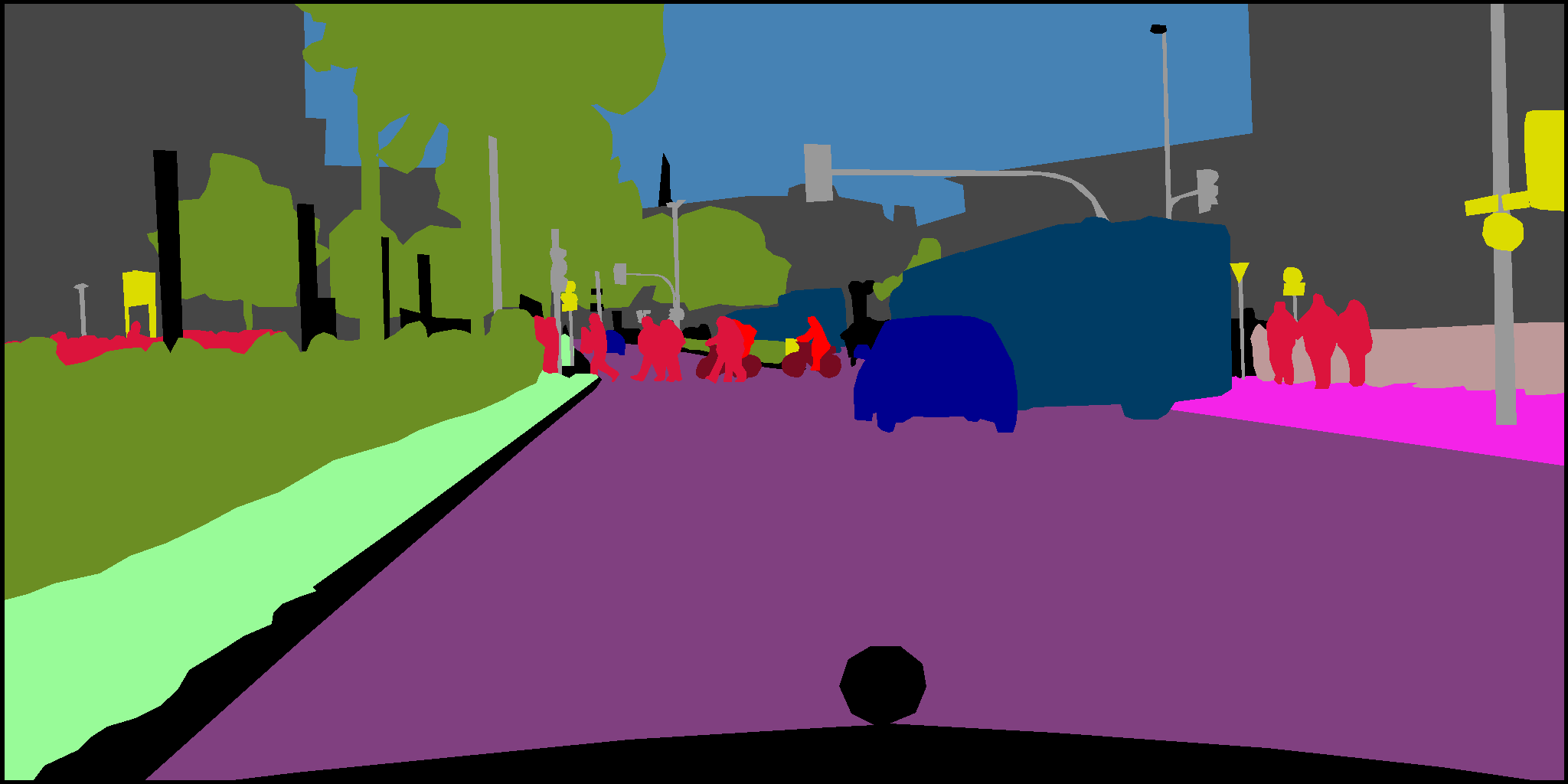}
	\includegraphics[width=.24\linewidth]{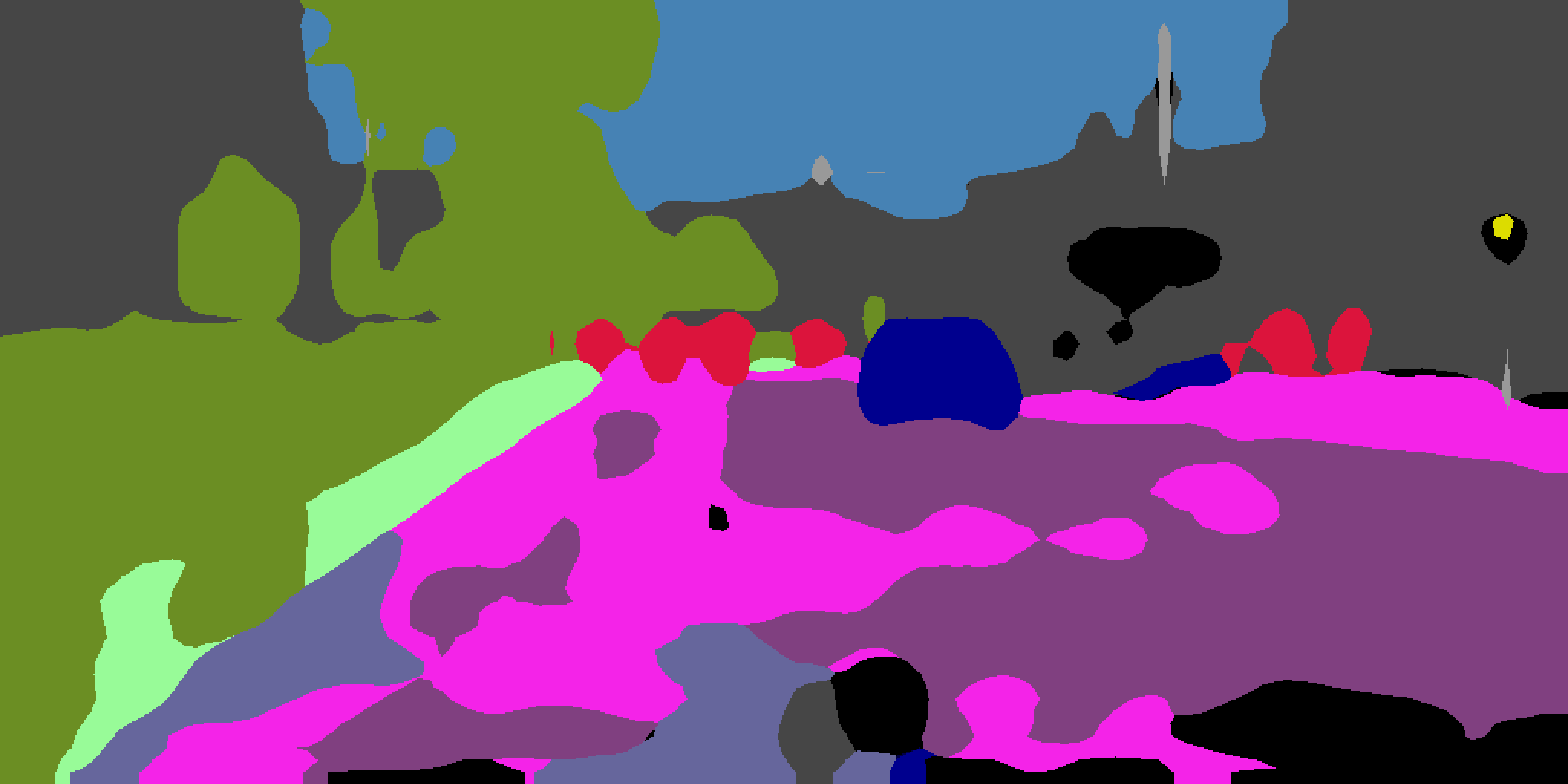}
	\includegraphics[width=.24\linewidth]{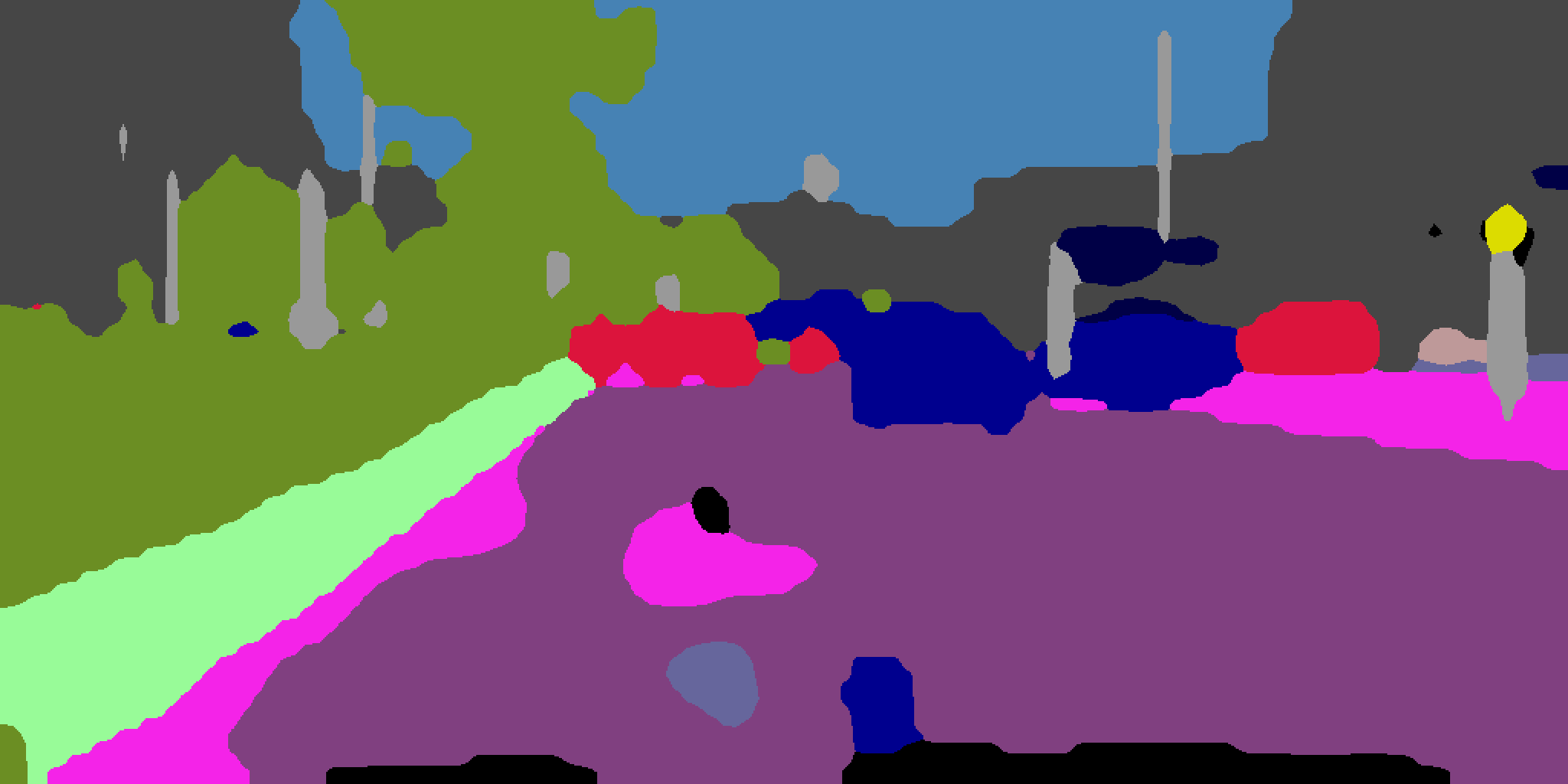}
	
	\includegraphics[width=.24\linewidth]{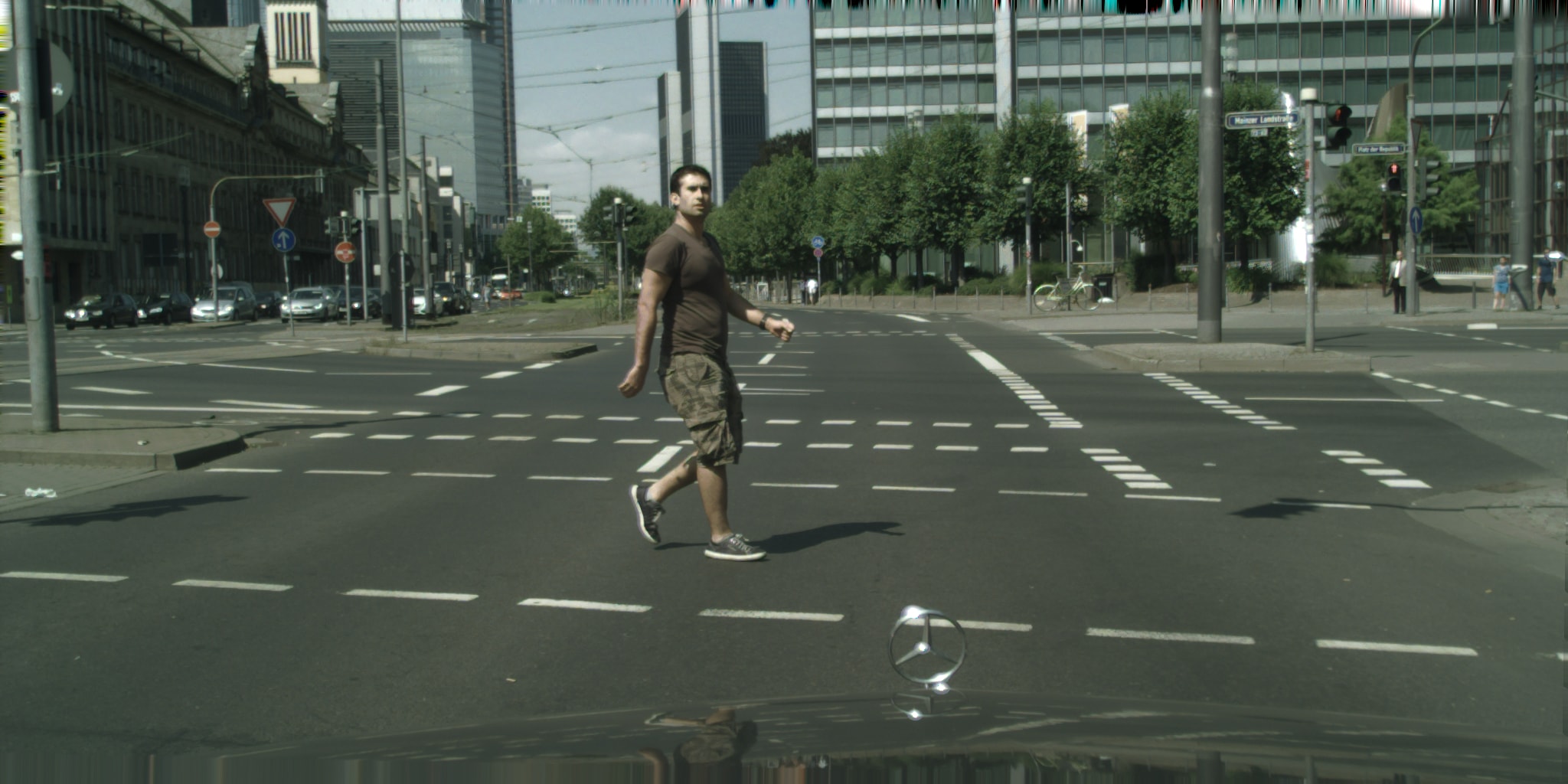}
	\includegraphics[width=.24\linewidth]{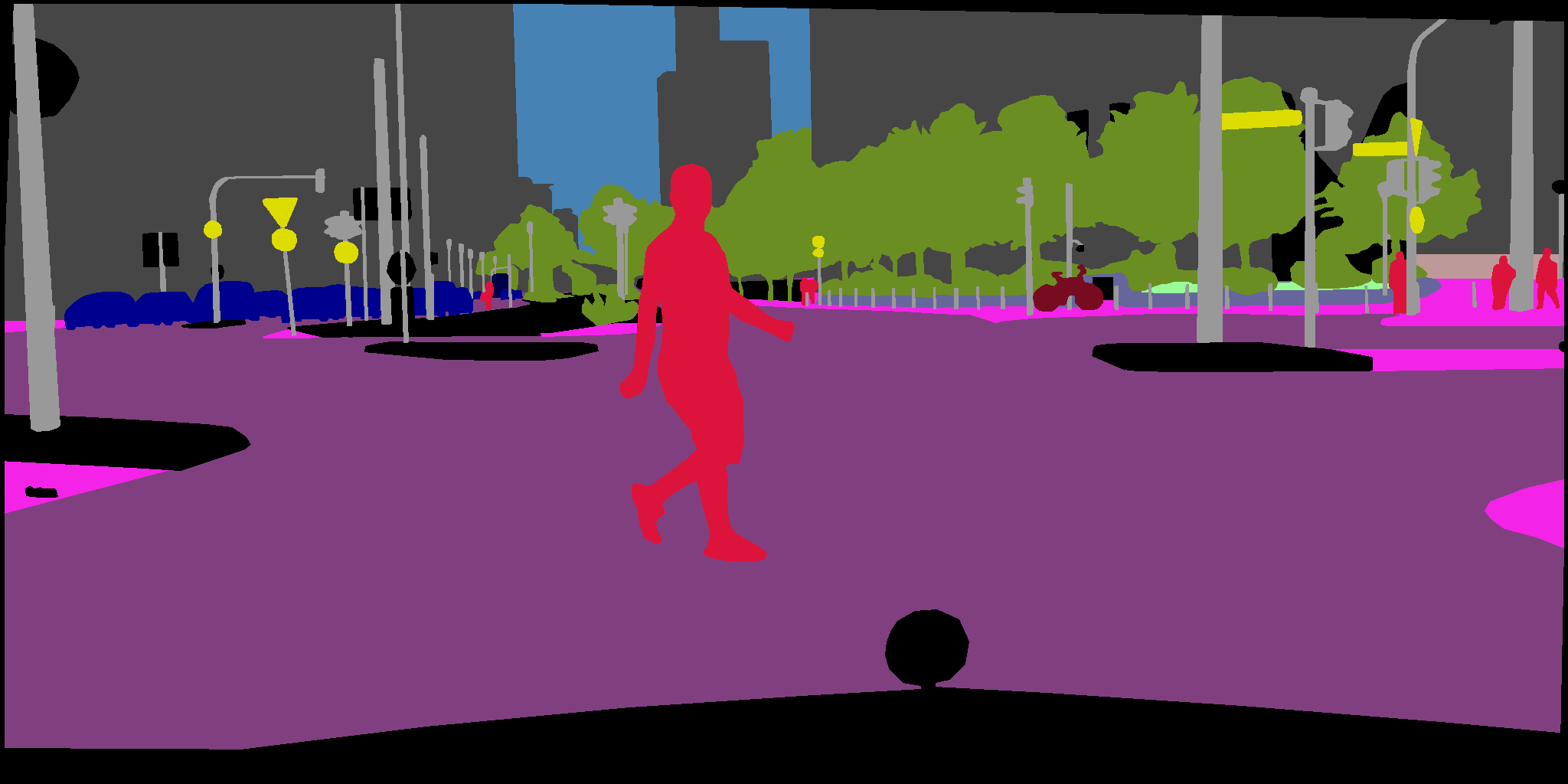}
	\includegraphics[width=.24\linewidth]{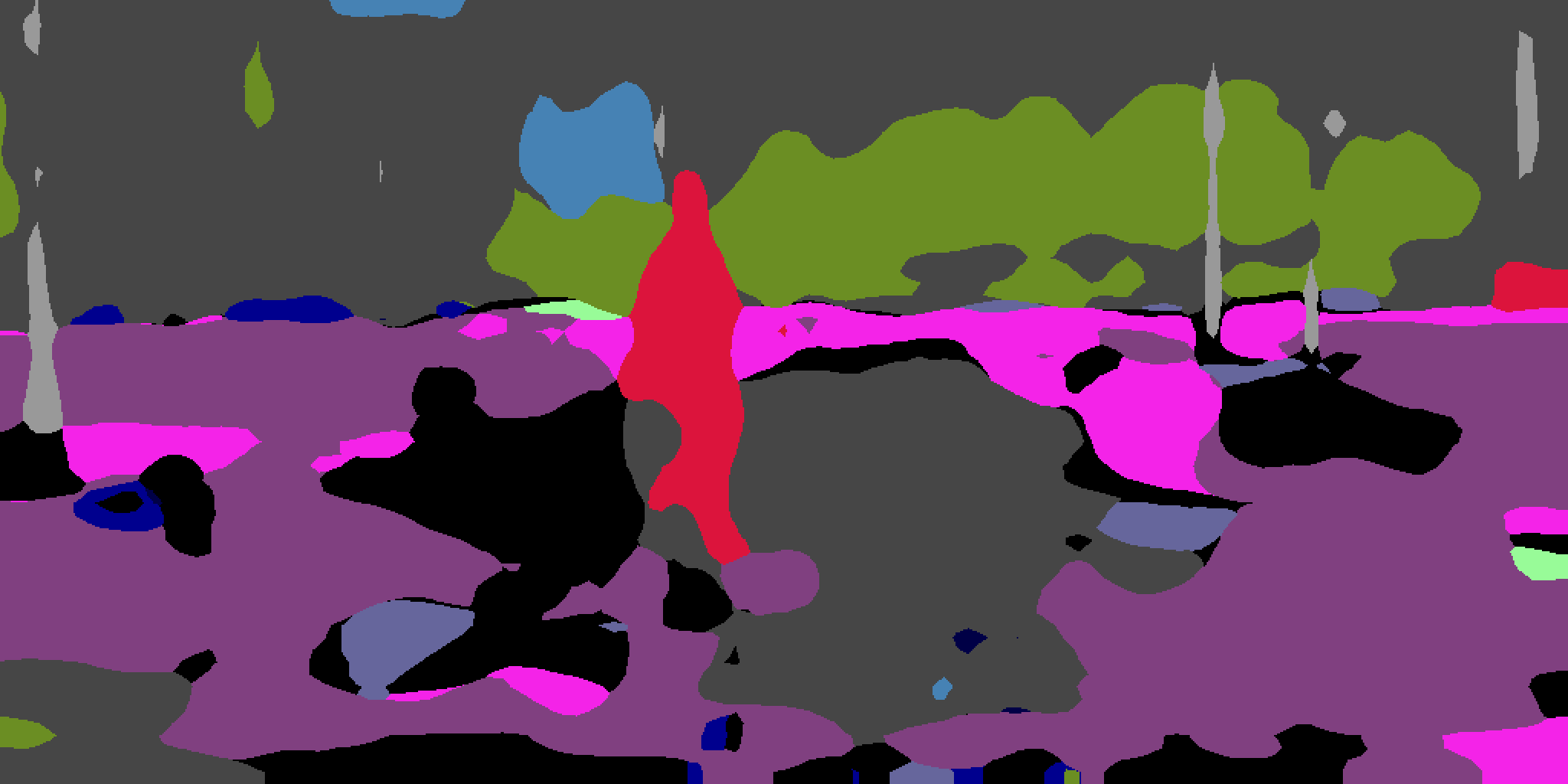}
	\includegraphics[width=.24\linewidth]{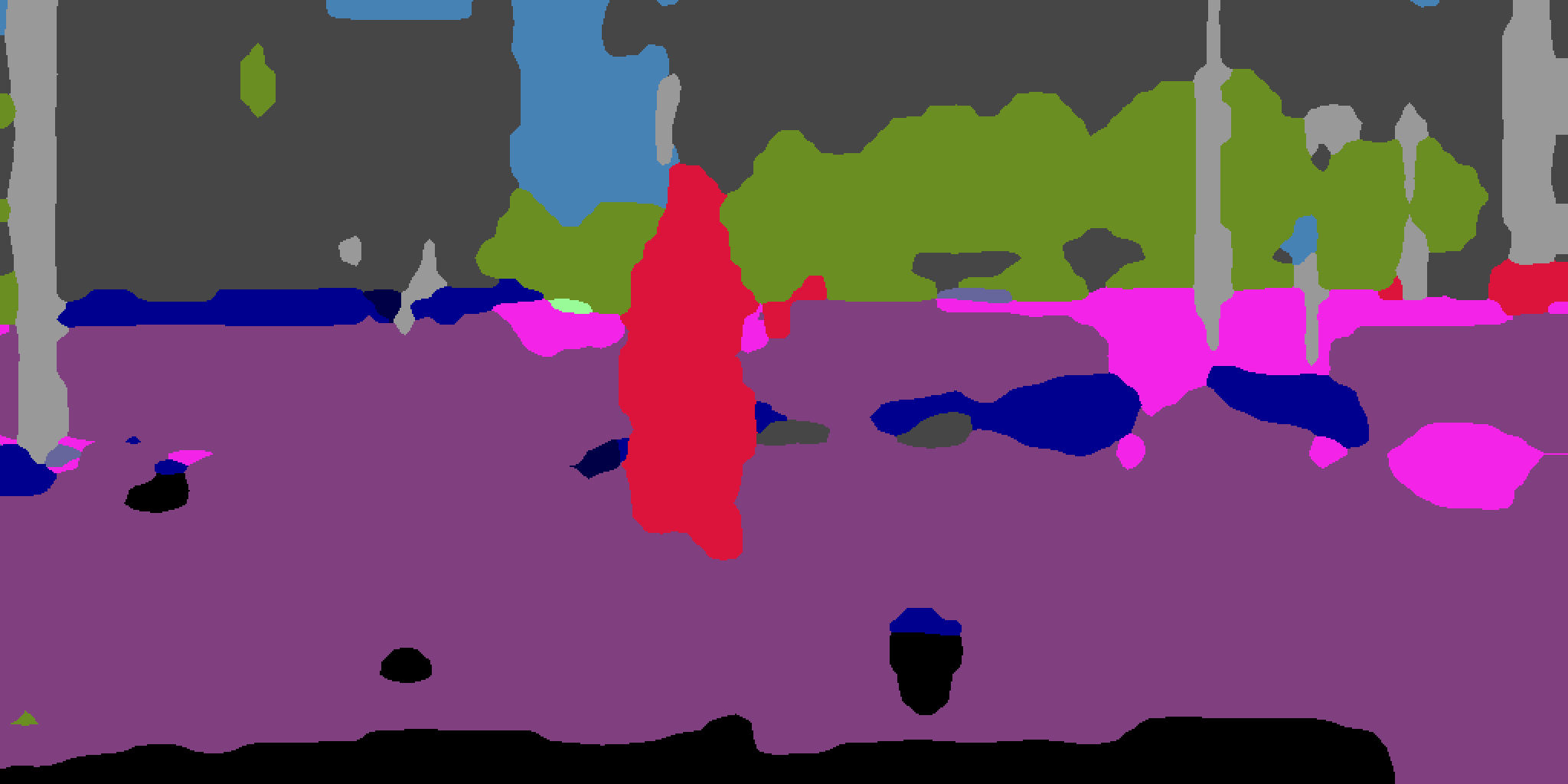}
	
	\includegraphics[width=.24\linewidth]{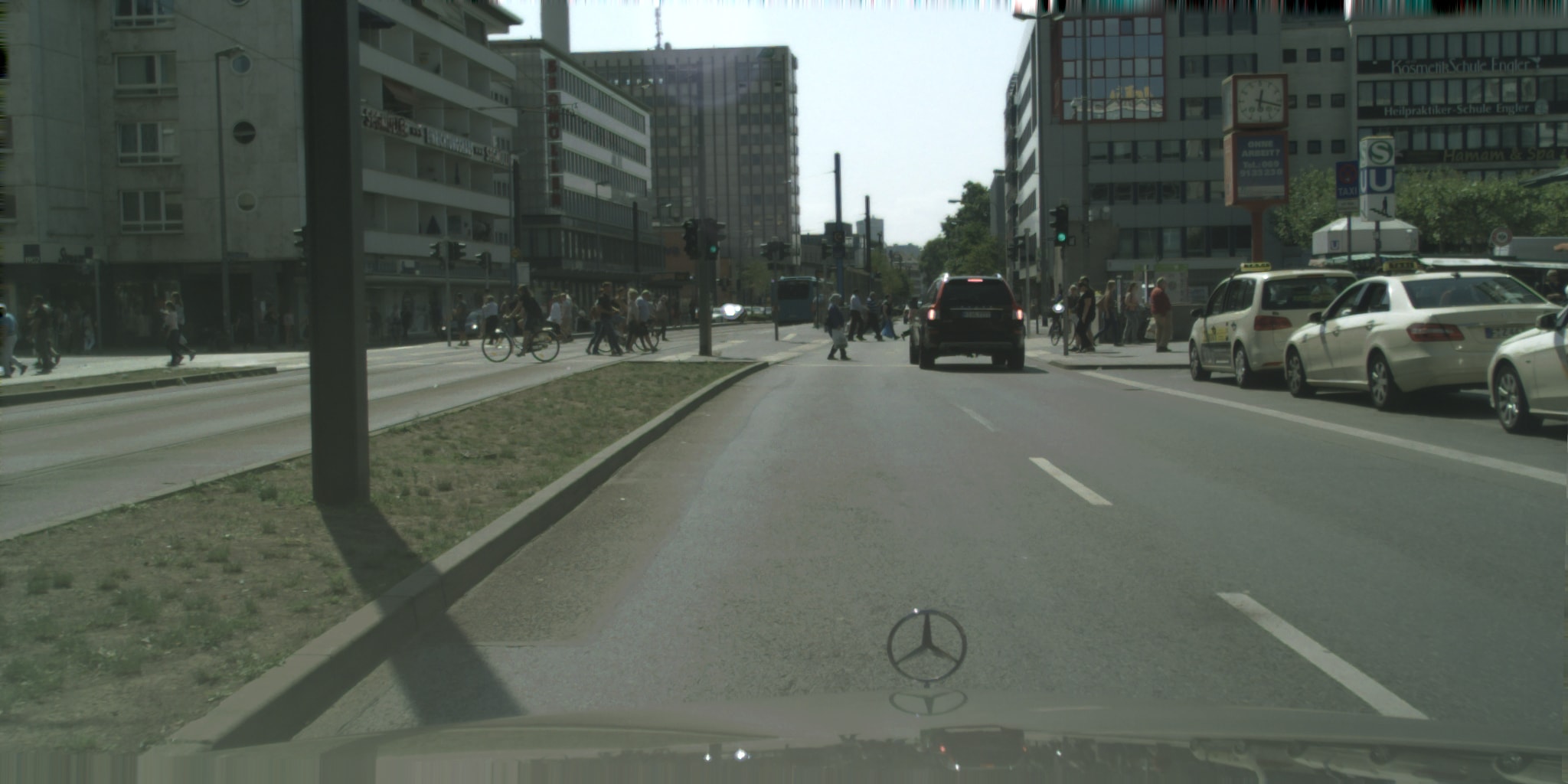}
	\includegraphics[width=.24\linewidth]{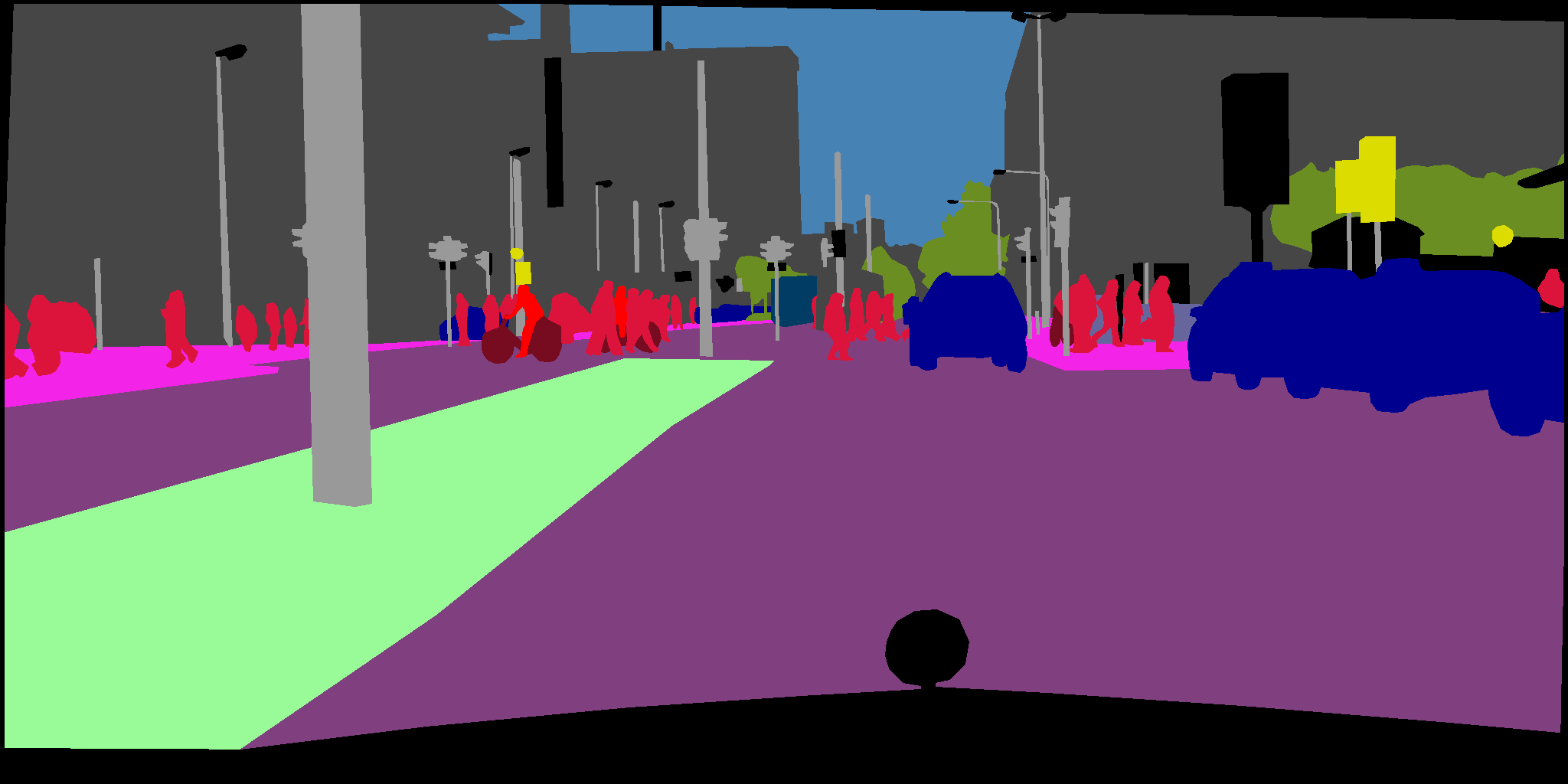}
	\includegraphics[width=.24\linewidth]{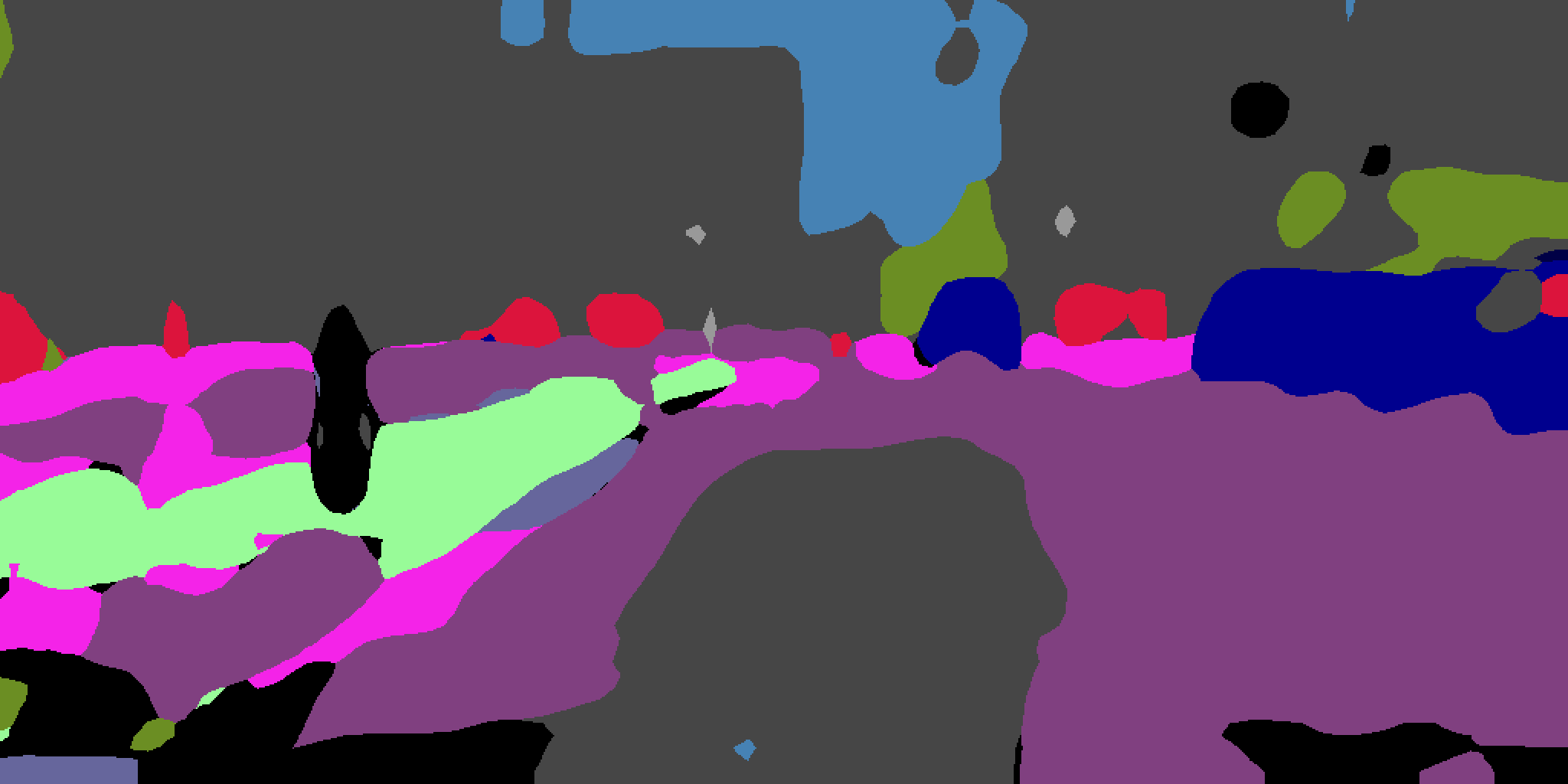}
	\includegraphics[width=.24\linewidth]{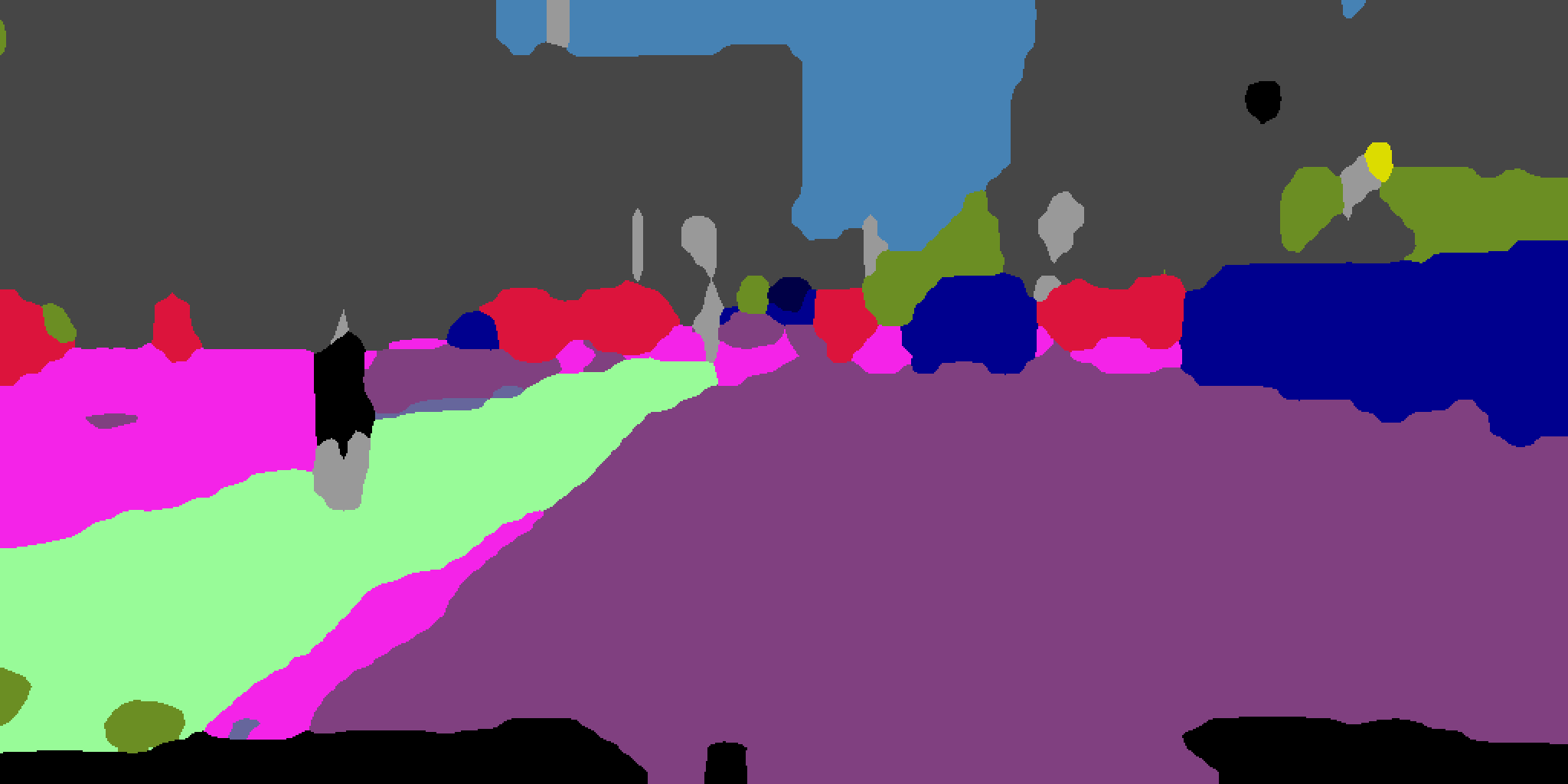}
	
	\begin{minipage}[b]{.24\linewidth}
		\centering
		\centerline{Input Image}\medskip
	\end{minipage}
	\begin{minipage}[b]{0.24\linewidth}
		\centering
		\centerline{}
		\centerline{Ground Truth}\medskip
	\end{minipage}
	\begin{minipage}[b]{0.24\linewidth}
		\centering
		\centerline{No Adapt}\medskip
	\end{minipage}
	\begin{minipage}[b]{0.24\linewidth}
		\centering
		\centerline{Ours}\medskip
	\end{minipage}
\caption{Domain adaptation results from the Cityscapes \texttt{Val} set. The
third column shows segmentation results using the model trained solely
by the GTA dataset, and the fourth column shows the segmentation results
after two rounds of the FCTN training (best viewed in color).}\label{fig:GTA_res}
\vspace{-1em}
\end{figure*}

\subsection{Training Procedure}\label{ssec:training}

The training process is illustrated in Algorithm \ref{alg:tri-training}.
We first pretrain the entire FCTN on the labeled source domain training
set $\mathcal{S}$ for $iters$ iterations, optimizing the loss function in
Eq. (\ref{eq:pretrain}). We then use the pre-trained model to generate
the initial pseudo labels for the target domain training set
$\mathcal{T}$, using the method described in Sec. \ref{ssec:labeling}.
We re-train the network using $\mathcal{S}$ and $\mathcal{T}_l$ for
several steps. At each step, we take a minibatch of samples with half
from $\mathcal{S}$ and half from $\mathcal{T}_l$, optimizing the terms
in Eq. (\ref{eq:curriculum}) jointly. We repeat the re-labeling of
$\mathcal{T}$ and the re-training of the network for several rounds
until the model converges. 

\begin{algorithm}[H]
	\caption{Training procedure for our fully convolutional tri-branch network (FCTN). }
	\label{alg:tri-training}
	\begin{algorithmic}
	\renewcommand{\algorithmicrequire}{\textbf{Input:}}
	\renewcommand{\algorithmicensure}{\textbf{Output:}}
	\Require labeled source domain training set $\mathcal{S} = \{(x_i^s, y_i^s)\}_{i=1}^{n_s}$ and unlabeled target domain training set $\mathcal{T} = \{x_i^t\}_{i=1}^{n_t}$
	\\ \textit{Pretraining on $\mathcal{S}$} :
	\For {$i = 1$ to $iters$}
	\State train $F, F_1, F_2, F_t$ with minibatches from $\mathcal{S}$
	\EndFor
	\\ \textit{Curriculum Learning with $\mathcal{S}$ and $\mathcal{T}$} :
	\For {$i = 1$ to $rounds$}
	\State $\mathcal{T}_l \gets$ \Call{Labeling}{$F, F_1, F_2, \mathcal{T}$} \Comment{See Sec. \ref{ssec:labeling}}
	\For {$k = 1$ to $steps$}
	\State train $F, F_1, F_2$ with samples from $\mathcal{S}$
	\State train $F, F_1, F_2, F_t$ with samples from $\mathcal{T}_l$
	\EndFor
	\EndFor\\
	\Return $F, F_t$
\end{algorithmic}
\end{algorithm}

\section{Experiments}\label{sec:experiments}

We validate the proposed method by experimenting the adaptation from the
recently built synthetic urban scene dataset GTA
\cite{richter2016playing} to the commonly used urban scene semantic
segmentation dataset Cityscapes \cite{Cordts2016Cityscapes}. 


Cityscapes \cite{Cordts2016Cityscapes} is a large-scale urban scene
semantic segmentation dataset. It provides over 5,000 finely labeled
images (train/validation/test: 2,993/503/1,531), which are labeled with
per pixel category labels. They are with high resolution of $1024 \times
2048$. There are 34 distinct semantic classes in the dataset, but only
19 classes are considered in the official evaluation protocol. 

GTA \cite{richter2016playing} contains 24,966 high-resolution labeled
frames extracted from realistic open-world computer games, Grand Theft
Auto V (GTA5). All the frames are vehicle-egocentric and the class
labels are fully compatible with Cityscapes. 

We implemented our method using Tensorflow\cite{abadi2016tensorflow} and
trained our model using a single NVIDIA TITAN X GPU. We initialized the
weights of shared base net $F$ using the weights of the
VGG-16 model pretrained on ImageNet. The hyper-parameter settings were
$\alpha=10^{3}, \beta=100$. We used a constant learning rate $10^{-5}$
in the training.  We trained the model for $70k$, $13k$ and $20k$
iterations in the pre-training and two rounds of curriculum learning,
respectively. 


We use synthetic data as source labeled training data and Cityscapes
\texttt{train} as an unlabeled target domain, while evaluating our
adaptation algorithm on Cityscapes \texttt{val} using the predictions
from the target specific branch $F_t$.  Following Cityscapes official
evaluation protocol, we evaluate our segmentation domain adaptation
results using the per-class intersection over union (IoU) and mean IoU
over the 19 classes. The detailed results are listed in Table.
\ref{tab:GTA} and some qualitative results are shown in Fig.
\ref{fig:GTA_res}. We achieve the state-of-the-art domain adaptation
performance. Our two rounds of curriculum learning boost the mean IoU
over our non-adapted baseline by 2.7\% and 4.3\%, respectively.
Especially, the IoU improvement for the small objects (\emph{e.g.} pole,
traffic light, traffic sign etc.) are significant (over 10\%). 

\section{Conclusion}\label{sec:conclusion}

A systematic way to address the unsupervised semantic segmentation
domain adaptation problem for urban scene images was presented in this
work. The FCTN architecture was proposed to generate high-quality pseudo
labels for the unlabeled target domain images and learn from pseudo
labels in a curriculum learning fashion. It was demonstrated by the DA
experiments from the large-scale synthetic dataset to the real image
dataset that our method outperforms previous benchmarking methods by a
significant margin. 

There are several possible future directions worth exploring.  First, it
is interesting to develop a better weight constraint for the two
labeling branches so that even better pseudo labels can be generated.
Second, we may impose the class distribution constraint on each
individual image \cite{Zhang_2017_ICCV} so as to alleviate the confusion
between some visually similar classes, \emph{e.g.} road and sidewalk,
vegetation and terrain etc. Third, we can extend the proposed method to
other tasks, \emph{e.g.} instance-aware semantic segmentation. 


\vfill\pagebreak

\bibliographystyle{IEEEbib}
\bibliography{WeeklyReportRef}

\end{document}